\documentclass[10pt,twocolumn,letterpaper]{article}

\usepackage[pagenumbers]{iccv} % To force page numbers, e.g. for an arXiv version

\definecolor{iccvblue}{rgb}{0.21,0.49,0.74}
\usepackage[pagebackref,breaklinks,colorlinks,allcolors=iccvblue]{hyperref}

\usepackage{graphicx}
\usepackage{booktabs}

\usepackage{amsmath}
\usepackage{amssymb}
\usepackage{multirow}
\usepackage{xspace}
\usepackage{paralist}
\usepackage{arydshln}
\usepackage{microtype}
\usepackage[font=small]{caption}
\usepackage{bm}
\usepackage{bbm}
\usepackage{tabularx}
\usepackage[normalem]{ulem}
\useunder{\uline}{\ul}{}
\usepackage{xcolor}         % colors
\usepackage{wrapfig}
\usepackage{nicefrac}       % compact symbols for 1/2, etc.
\usepackage{amsmath,amssymb} 
\usepackage{comment}
\usepackage[utf8]{inputenc} % allow utf-8 input
\usepackage[T1]{fontenc}    % use 8-bit T1 fonts
\usepackage{url}            % simple URL typesetting
\usepackage{adjustbox}
\usepackage{xkeyval}
\usepackage{tikz}
\usepackage{calc}
\usepackage{colortbl}

\def\paperID{6715} % *** Enter the Paper ID here
\def\confName{ICCV}
\def\confYear{2025}

\usepackage{orcidlink}
\def\etal{\emph{et al}.~}
\def\eg{\emph{e}.\emph{g}.}
\def\ie{\emph{i}.\emph{e}.}

\usepackage[capitalize]{cleveref}
\crefname{section}{Sec.}{Secs.}
\Crefname{section}{Section}{Sections}
\Crefname{table}{Table}{Tables}
\crefname{table}{Tab.}{Tabs.}

\newcommand{\mysection}[1]{\vspace{0mm}\section{#1}\vspace{0mm}}
\newcommand{\mysubsection}[1]{\vspace{-0.5mm}\subsection{#1}\vspace{-0.3mm}}
\newcommand{\mypar}[1]{\vspace{-1.5mm}\paragraph{#1}}

\newcommand{\convabb}{GMR\xspace}
\newcommand{\convname}{GMR-Conv\xspace}

\title{\convname: An Efficient Rotation and Reflection Equivariant \\Convolution Kernel Using Gaussian Mixture Rings}

\author{
Yuexi~Du$^{1}$,~
Jiazhen~Zhang$^{1}$,~
Nicha~C.~Dvornek$^{1,2}$,~ 
John~A.~Onofrey$^{1,2,3}$\\
Department of Biomedical Engineering\\
Department of Radiology \& Biomedical Imaging,~Department of Urology, \\Yale University, New Haven, CT, USA \\
{\tt\small\{yuexi.du,jiazhen.zhang, nicha.dvornek,john.onofrey\}@yale.edu}
}

\begin{document}
\maketitle

%%%%%%%%% ABSTRACT
\begin{abstract}
    Symmetry, where certain features remain invariant under geometric transformations, can often serve as a powerful prior in designing convolutional neural networks (CNNs). While conventional CNNs inherently support translational equivariance, extending this property to rotation and reflection has proven challenging, often forcing a compromise between \textbf{equivariance}, \textbf{efficiency}, and \textbf{information loss}. In this work, we introduce Gaussian Mixture Ring Convolution (\convname), an efficient convolution kernel that smooths radial symmetry using a mixture of Gaussian-weighted rings. This design mitigates discretization errors of circular kernels, thereby preserving robust rotation and reflection equivariance without incurring computational overhead. 
    We further optimize both the space and speed efficiency of \convname via a novel parameterization and computation strategy, allowing larger kernels at an acceptable cost.
    Extensive experiments on eight classification and one segmentation datasets demonstrate that \convname not only matches conventional CNNs' performance but can also surpass it in applications with orientation-less data. \convname is also proven to be more robust and efficient than the state-of-the-art equivariant learning methods. Our work provides inspiring empirical evidence that carefully applied radial symmetry can alleviate the challenges of information loss, marking a promising advance in equivariant network architectures. 
    \footnote{Code is available at \url{https://github.com/XYPB/GMR-Conv}.}
\end{abstract}

%%%%%%%%% BODY TEXT
\mysection{Introduction}
\label{sec:intro}

Symmetry refers to a property where certain aspects or features remain unchanged (invariant) when specific transformations are applied, \eg, pathological images' structures are symmetrical under rotation. It is often beneficial to encode the prior knowledge of symmetry in the model if the data follows the symmetry, \eg, conventional convolutional neural networks (CNNs) are designed to be equivariant under translational transform so that it can capture the same feature at any position of the input. 
The ability to preserve the structure of transformations in the data and maintain symmetry is known as \emph{equivariance}. 
This ability is difficult to learn directly via data augmentation, as global augmentation does not cover all situations. 

\begin{figure*}[!t]
    \centering
    \includegraphics[width=\linewidth]{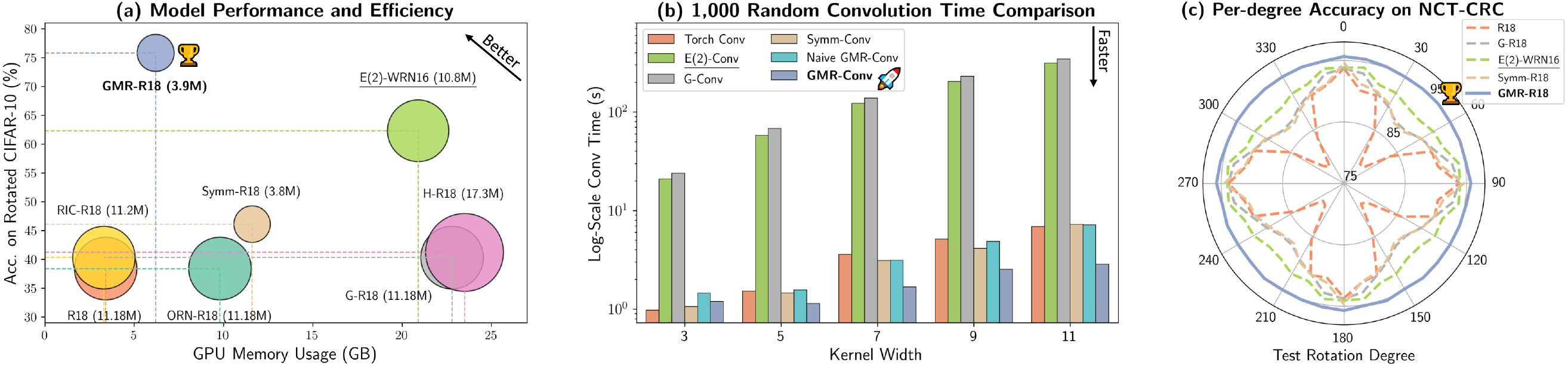}
    \vspace{-6mm}
    \caption{{\bf Advantages of \convname}. 
    \textbf{(a) High performance with low complexity:} Model size, GPU memory, and performance on rotated CIFAR-10 test set. The size of each circle reflects the model size. 
    \textbf{(b) Efficient computation:} Time of 1,000 random convolution operations at different convolutional kernel widths. 
    \textbf{(c) Superior equivariance:} Per-degree performance on orientation-less NCT-CRC dataset of pathological images. The SoTA method is underlined, and ours is highlighted in bold.} 
    \vspace{-4mm}
    \label{fig:teaser}
\end{figure*}  

Rotational and reflective equivariance is gaining more attention~\cite{basheer2024current} as it is a desired property for data with no explicit orientation~\cite{veeling2018rotation,fei2024rotation,he2024soe,zhu2024sre}, such as remote sensing images~\cite{wu2020benchmark,su2019object} or pathological images~\cite{kather2019predicting,bejnordi2017diagnostic,mahbod2021cryonuseg}. However, resolving the ``trilemma'' of (1) \textbf{Equivariance}, (2) \textbf{Efficiency}, and (3) \textbf{Avoiding Information Loss} all at once is a major challenge in the field. 
As the state-of-the-art in equivariant learning, the grouped steerable filter-based method~\cite{weiler2018learning,weiler2019general,cesa2021program,cohen2016group,worrall2017harmonic,karella2024harmformer} is lossless and maintains rotational and reflective equivariance under a pre-defined basis, \eg, $\{\frac{\pi}{2}, \pi, \frac{3\pi}{2}, 2\pi\}$ (G-R18 and E(2)-Conv in \cref{fig:teaser}(c)). However, these methods demand larger model sizes, higher computational costs, and are inefficient (\cref{fig:teaser}(a, b)). 
Another group of solutions uses invariant coordinate systems, \eg, log-polar coordinates, but this results in loss of the phase information~\cite{esteves2017polar, kim2020cycnn, paletta2022spin, kim2020cycnn} or loss of translational equivariance~\cite{mo2022ric}, which naturally leads to lower performance (RIC-R18 in \cref{fig:teaser}(a)).
Radial symmetric convolution~\cite{yeh2016stable,kohli2017learning,dudar2019use,fuhl2021rotated,zhang2019shellnet}, an intuitive method for equivariance~\cite{Marr1980-pn}, has received less attention despite its advantages of no extra computation and greater efficiency compared with other methods. 
A noted limitation of the radial symmetric convolution is that it cannot realize true equivariance in the discrete setting (Symm-R18 in \cref{fig:teaser}(c)) despite it being perfectly rotation and reflection-equivariant in continuous space (\cref{sec:sre}). 
The discretization of small, square convolutional kernels violates the radial symmetry, resulting in a lack of equivariance under rotation as discussed in \cref{sec:kernel_config}. Furthermore, the symmetry constraint makes the kernel ``invariant'' under local rotation, and therefore leads to information loss and suboptimal performance.

In this paper, we extend and enhance the design of radial symmetric convolution kernels to overcome their known limitations. Our approach proposes an efficient rotation and reflection equivariant convolution kernel using Gaussian Mixture Rings (\convabb). Our experiments on 8 classification~\cite{krizhevsky2009learning,su2019object,wu2020benchmark,kather2019predicting,bejnordi2017diagnostic,li2009imagenet,wu20153d} and 1 segmentation~\cite{mahbod2021cryonuseg} datasets demonstrate that our \convname achieves the best balance for the ``trilemma''.
Extending the design of an efficient radial symmetric convolution kernel, we propose to use a mixture of rings with a width defined by a Gaussian distribution to \emph{smooth} the kernel and avoid equivariance errors caused by discretization (\cref{fig:teaser}(c)). We further propose a highly-efficient (memory and computational) kernel parametrization and computation strategy to achieve the best efficiency in terms of both size and speed (\cref{fig:teaser}(a, b)), allowing us to use a larger convolution kernel while maintaining efficiency. While our method does not explicitly resolve the problem of information loss, our observation shows that our symmetry \convname can match the performance of conventional CNN and even outperform in specific datasets with orientation-less data (\cref{tab:main}), despite its radial symmetry constraint. Our results on the nuclei segmentation~\cite{mahbod2021cryonuseg} task also demonstrate its capability in more complex tasks (\cref{tab:seg}). The plug-and-play nature of \convname allows it to be deployed within any CNN architecture to achieve rotational and reflection equivariance and seamlessly improve efficiency. Our contributions can be summarized as follows:
\begin{compactitem}
    \item[(1)] \textbf{Equivariance:} We propose \convname that uses Gaussian Mixture Rings to smooth the radial symmetric kernel, achieving the best rotation and reflection equivariance across multiple datasets (\cref{tab:main}).
    \item[(2)] \textbf{Efficiency:} We propose an efficient parametrization method and a novel computational strategy to implement our radial symmetric \convname, achieving the optimized memory usage and highest speed efficiency (\cref{fig:teaser} (a, b)), allowing larger convolution kernels at an acceptable cost.
    \item[(3)] \textbf{Information Loss:} We provide a novel observation that a strictly radial symmetric convolution kernel with a large receptive field can match the performance of a conventional convolution layer and even perform better in specific data like remote sensing and pathological images, alleviating concerns of information loss.
\end{compactitem}

\mysection{Related Work}

\subsection{Rotation-Encoded Neural Networks} Neural networks that encode a fixed number of pre-defined rotations are a common approach to achieving rotational equivariance. H-Net~\cite{worrall2017harmonic} and Steerable Filter CNN~\cite{weiler2018learning,weiler2019general,cesa2021program} make use of the property of circular harmonic or steerable filters to encode rotation in specific angles to the model and disentangle the input to these pre-defined bases. Dieleman~\etal\cite{dieleman2016exploiting} proposed a cyclic operation network to encode the features with different orientations in the batch and channel dimensions. Another method involves rotating the convolutional filters. Inspired by the group-equivalent operations, G-CNN~\cite{cohen2016group} rotates and reflects the convolutional filters every $90^\circ$ to ensure rotational equivariance. In ORN~\cite{zhou2017oriented}, Active Rotating Filters will be rotated during convolution to encode feature maps in different orientations. 
Similar attempts have been made to rotate the filters and gain a rotation-equivariant property~\cite{linmans2018sample, chidester2019rotation}. 
Follmann~\etal\cite{follmann2018rotationally} rotates the feature map obtained by the rotated filter to embed the feature in four different orientations. Alternatively, other works process inputs of different orientations to make networks aware of rotational equivariance~\cite{cabrera2017deep, zhou2022mtcnet, YAO202222, gupta2021rotation}. Karella~\etal\cite{karella2024harmformer} has integrated such a design into a transformer. These methods are theoretically equivalent to the methods that rotate the filters. However, these methods bring excessive size and computational cost as the number of pre-defined angles increases. Meanwhile, these methods show poor performance for the angles that are not pre-defined.

\subsection{Rotation-Invariant Coordinate Systems} Another strategy is to transform the input data to a different coordinate system to ensure rotation equivariance. Mo~\etal\cite{mo2022ric} uses a cyclic rotation-invariant coordinate system to ensure that convolutions follow a specific sequence regardless of input rotation. However, this method cannot guarantee translational equivariance of the convolution operation. Other attempts include transferring the input data to polar or log-polar systems~\cite{esteves2017polar, kim2020cycnn, paletta2022spin}. These methods benefit from the property of the polar coordinate system, in which translation is equivalent to rotation in the Cartesian coordinate system. However, the polar mapping will naturally result in the loss of the phase information, and the image will also be distorted. PTN~\cite{esteves2017polar} addresses this issue by introducing a network to predict the center of polar mapping, while CyCNN~\cite{kim2020cycnn} wraps the mapped image into a cylindrical shape and slides the convolutional window over it. Despite this improvement, methods using polar mapping may still result in information loss.

\subsection{Radial Symmetric Convolution} Radial symmetric convolution is one of the most intuitive solutions to achieve rotation equivariance. Kohli~\cite{kohli2017learning}~\etal was one of the first to apply a radial symmetric convolutional kernel to equivariance learning.
SymNet~\cite{dzhezyan2019symnet} proposes a horizontal symmetric convolutional kernel to make the model more robust to horizontal reflections. Yeh~\etal\cite{yeh2016stable} and Dudar~\etal\cite{dudar2019use} explore using different shapes of the symmetric convolutional kernel. Dudar~\etal\cite{dudar2019use} and Fuhl~\etal\cite{fuhl2021rotated} first explore the potential of using a symmetric convolution kernel to achieve rotational equivariance. ShellNet~\cite{zhang2019shellnet} explores radial symmetry kernel in 3D object recognition. However, these existing methods construct the radial symmetric kernel based on its radius to the center, resulting in a strong discretization error. The trivial computational strategy further constrains larger convolutional kernels, leading to suboptimal performance due to information loss.

\begin{figure*}[t!]
    \centering
    \includegraphics[width=\textwidth]{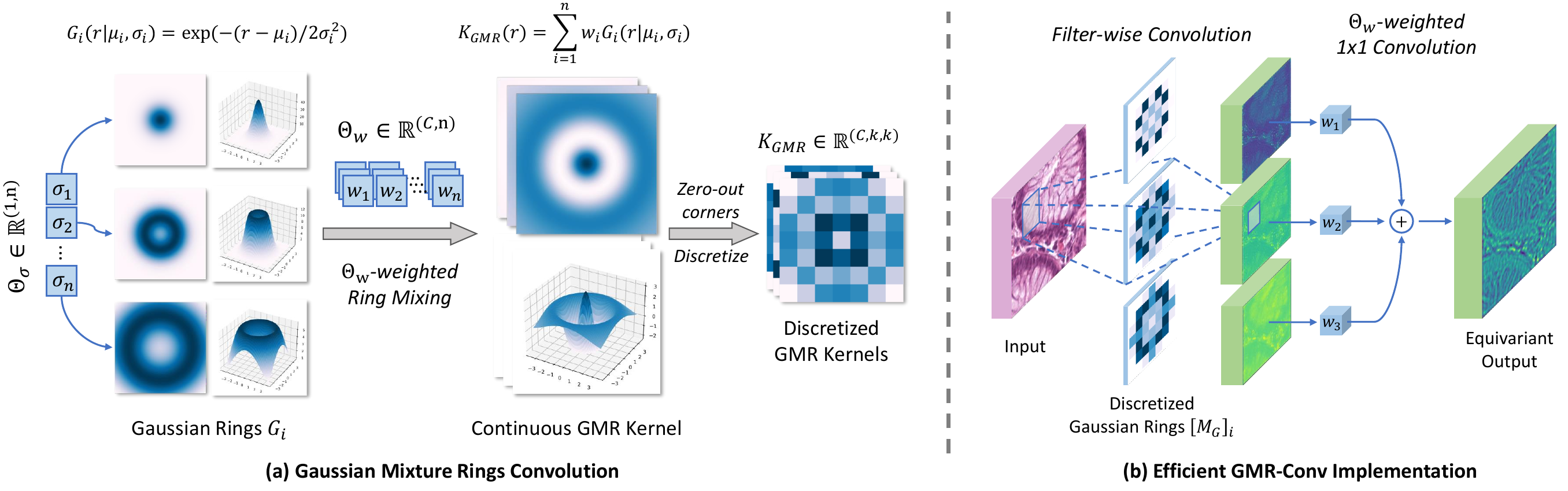}
    \caption{{\bf Building the \convname Kernel Efficiently}. (a) We build the \convname kernel from the Gaussian rings controlled by $\Theta_\sigma$ and weighted by $\Theta_w$. The final kernel will be a weighted sum of each ring. (b) To convolve the proposed \convname kernel efficiently, we split it into a per-Gaussian ring depthwise convolution and a $\Theta_w$ weighted 1-by-1 convolution to reduce computational complexity.
    }  
    \vspace{-1mm}
    \label{fig:build}
    \vspace{-3mm}
\end{figure*}  
\mysection{Methods}

Here, we introduce how to build our equivariant and symmetric Gaussian Mixture Ring (\convabb)  convolution (\convname) kernel and a novel parameter and computational-efficient implementation of \convname.

\mysubsection{Symmetric Rotation Equivariant Kernels}
\label{sec:sre}

We approach rotational equivariance with a centrally symmetric kernel~\cite{Marr1980-pn,dudar2019use}, where we parameterize the 2D kernel using independent circular rings according to their Euclidean distance to the center of the kernel. Here, each ring corresponds to one trainable parameter. Namely, all the kernel parameters symmetric to the center of the kernel will share the same value. Such a centrally symmetric kernel achieves \emph{local} rotational and reflection invariance with respect to the Hadamard product and \emph{global} rotational and reflection equivariance under the convolution operation as the kernel slides through the input. Here, we provide proof of equivariance in continuous space in the 2D case without loss of generality. Given a 2D function $f(x, y)$, we define the rotation operation with respect to the origin as $R(\cdot)$. The rotation equivariance is defined as $R(F[f(x, y)]) = F[R(f(x, y))]$, where $F$ is a function of a function. Consider a central symmetric function $h(x,y)$. The rotation of the convolution of $h$ and $f$ is:
\begin{align*}
    R(h*f)(x,y) &= R(\iint h(u,v)f(x-u, y-v)dudv)\\
    &=\iint h(u',v')f(x'-u', y'-v')du'dv',
\end{align*}
where $(x', y')=(x\cos\theta-y\sin\theta, x\sin\theta+y\cos\theta)$ and same for $(u', v')$. We know $f(x'-u', y'-v')=R(f(x-u,y-v))$ since rotation is a linear operation, \ie, $x'-u' = (x-u)'$. We then compute the Jacobian: 
\begin{align*}
    du'dv'&=(\frac{\partial u'}{\partial u}\frac{\partial v'}{\partial v} -\frac{\partial v'}{\partial u}\frac{\partial u'}{\partial v})dudv \\
    &=(\cos^2\theta+\sin^2\theta)dudv=dudv.
\end{align*}
Thus, the rotation of the convolution of $f$ and $h$ is:
\begin{align*}
    R(h*f)(x, y)&=\iint R(h(u,v))R(f(x-u, y-v))dudv\\
    &=R(h)*R(f)=h*R(f)
\end{align*}
since $h$ is centrally symmetric and is naturally rotation-invariant. This completes our proof of rotation \emph{equivariance}.

Benefiting from the translation equivariant nature of the convolution operation, the local rotation and reflection invariant kernel can generate equivariant global output. This property also ensures that the kernel is not only equivariant to the rotation of the whole input but also equivariant to sub-region rotation and reflection.

\mysubsection{Efficient \convname Implementation}
\label{sec:gmr}

While the symmetric kernel can maintain perfect rotational and reflection equivariance under continuous space, as proved above, this property will be harmed by discretization. The effect of discretization is exacerbated when using a small kernel size like $3\times3$.
While using a larger kernel size can help mitigate this problem, it will also increase the number of parameters and slow down the computation. To alleviate the influence of discretization, we propose a Gaussian Mixture Ring (\convabb) (\cref{fig:build}(a)) method to smooth each independent ring and improve the efficiency. The bell-shaped distribution of the Gaussian function allows a smooth mixing only within neighboring rings and does not influence others.

Given a 2D kernel of size $k$-by-$k$, we split it into $n$ individual rings with equal width starting from the center. Each ring corresponds to one trainable parameter $w_i$ and a Gaussian function $G_i(r|\mu_i,\sigma_i)=\exp(-(r-\mu_i)/2\sigma_i^2)$, where $\mu_i$ and $\sigma_i$ is the mean and standard deviation respectively, and $r$ is the radius to the center of the kernel. 
The Gaussian function $G_i$ acts as an interpolating weight to smooth the values between each ring, where $\sigma_i$ controls the degree of smoothness.
Namely, the final 2D kernel $K_{\convabb}$ can be viewed as a weighted sum of $n$ rings parameterized with $w_i$
\begin{equation}
    K_{\convabb}(r) = \sum^n_{i=1} w_i G_i(r|\mu_i,\sigma_i),~r\in[0,\frac{k}{2}).
\end{equation}
We set $\mu_i=(i-1)\Delta d$, where $\Delta d=\frac{k}{2(n-1)}$ is the width of each ring. Similarly, we initialize $\sigma_i$ so that the full-width half max (FWHM) of $G_i$ equals $\Delta d$. We make the standard deviation $\sigma_i$ trainable so that the model has control over the width of each Gaussian ring and can adjust accordingly.

While the kernel space of $K_{\convabb}$ is the same as a discrete symmetric kernel, \convabb's smoothing forces the kernel to better approximate a continuous function and helps to mitigate errors due to discretization. This key design improves the performance on rotation angles where the discrete kernel is not strictly symmetric  (see ablation experiments). Despite our design not explicitly resolving the information loss issue, we empirically prove that symmetric \convname is capable of various complex data and tasks in the experiment. A more thorough discussion is provided in \cref{sec:tradeoff}.

\noindent\textbf{Discrete Implementation.}
For the kernel $K_{\convabb}\in\mathbb{R}^{(C, k, k)}$ ($C$ represents both $C_{in}$ and $C_{out}$ for simplicity) with $n$ rings, we use matrix $\Theta_w\in\mathbb{R}^{(C, n)}$ to represent the trainable parameters of each ring. Since there are $k^2$ positions in the kernel, each Gaussian ring $G_i$ can be discretized into $k^2$ elements. This results in a matrix $M_G\in\mathbb{R}^{(n, k, k)}$ representing the Gaussian weight for each ring:
\begin{equation}
    [M_G]_{i, u, v} = G_i(\sqrt{u^2+v^2}|\mu_i,\sigma_i),
\end{equation}
where $i\in\{1,\dots,n\}$, $u,v\in\{-\frac{k-1}{2}, \dots, 0, \dots, \frac{k-1}{2}\}$, and $\sigma_i$ is parameterized with $\Theta_{\sigma}\in\mathbb{R}^n$. Thus, the final \convname kernel $K_{\convabb}\in\mathbb{R}^{(C, k, k)}$ is given by:
\begin{equation}
    [K_{\convabb}]_{c,u,v}=\sum_{i=1}^n [\Theta_w]_{c,i}[M_G]_{i,u,v}
    \label{eq:discrete_kernel}
\end{equation}
This formulation allows us to easily extend \convname to a higher dimension while maintaining equivariance.

\noindent\textbf{Circular Kernel Constraint.} We note that rotated input performance degradation largely occurs for non-$90^\circ$ angles. To address this issue, we force the parameters at the 4 corners of the $K_{\convabb}$ kernel to be zero and ensure the kernel is closer to circular so that the kernel performs more consistently for non-$90^\circ$ angles during test time. 
Excluding the kernel corners further improves model efficiency.

\begin{table*}[!t]
\centering

\setlength{\tabcolsep}{8pt}
\resizebox{\textwidth}{!}
{
\begin{tabular}{lccccc|ccc|ccc|ccc|ccc}
\toprule
\multicolumn{1}{c}{\multirow{2}{*}{\textbf{Models}}} & \multicolumn{1}{c}{\multirow{2}{*}{\textbf{\#Param.}}} & \multicolumn{1}{c}{\multirow{2}{*}{\textbf{Memo.}}} & \multicolumn{3}{c|}{\textbf{CIFAR-10}~\cite{krizhevsky2009learning}} & \multicolumn{3}{c|}{\textbf{NWPU-10}~\cite{su2019object}} & \multicolumn{3}{c|}{\textbf{MTARSI}~\cite{wu2020benchmark}} & \multicolumn{3}{c|}{\textbf{NCT-CRC}~\cite{kather2019predicting}} & \multicolumn{3}{c}{\textbf{Patch-Camelyon}~\cite{bejnordi2017diagnostic}} \\ \cmidrule(l){4-18} 
\multicolumn{1}{c}{} & \multicolumn{1}{c}{} & \multicolumn{1}{c}{} & Orig. & Rot. & Ref. & Orig. & Rot. & Ref. & Orig. & Rot. & Ref. & Orig. & Rot. & Ref. & Orig. & Rot. & Ref. \\ \midrule
R18~\cite{he2016deep} & 11.2M & 3.4G & 87.7 & 38.4 & 64.2 & 97.6 & 90.9 & 95.3 & 92.7 & 57.2 & 65.4 & 93.7 & 87.3 & 92.9 & 84.8 & 75.3 & 82.6\\ \midrule
ORN-R18~\cite{zhou2017oriented} & 11.2M & 9.8G & 88.0 & 38.4 & 61.4 & {\ul 97.9} & 91.4 & 96.7 & 93.6 & 87.1 & 83.2 & 90.9 & 91.0 & 85.5 & 85.5 & 72.0 & 82.6\\
G-R18~\cite{cohen2016group} & 11.2M & 22.8G & {\ul 94.1} & 40.3 & 62.5 & 95.9 & 91.8 & 95.8 & {\ul 96.4} & 79.2 & 94.3 & 93.7 & 90.8 & 91.7 & \textbf{86.4} & 78.7 & {\ul 85.4} \\
H-R18~\cite{worrall2017harmonic} & 17.3M & 23.5G & 86.6 & 41.2 & 61.0 & 93.9 & 85.9 & 91.3 & 96.4 & 74.4 & 94.1 & 89.1 & 88.7 & 89.1 & 77.8 & 73.5 & 76.3\\
E(2)-WRN16~\cite{weiler2019general} & 10.8M & 20.9G & \textbf{94.4} & {\ul 62.3} & {\ul 84.9} & \textbf{98.3} & {\ul 96.3} & \textbf{97.7} & \textbf{96.7} & {\ul 93.0} & {\ul 95.6} & 93.8 & {\ul 92.5} & {\ul 93.9} & 85.4 & {\ul 82.6} & 85.1\\
~~-~$0^{th}$-order filter & 2.9M & 19.7G & 75.7 & 56.8 & 68.2 & 93.6 & 92.2 & 93.8 & 84.7 & 84.6 & 84.7 & 89.7 & 88.3 & 89.5 & 76.7 & 69.8 & 75.7 \\ 
RIC-R18~\cite{mo2022ric} & 11.2M & 3.3G & 87.4 & 40.3 & 70.7 & 96.2 & 92.4 & 95.2 & 89.4 & 63.7 & 87.7 & 90.6 & 91.0 & 90.7& 84.5 & 67.0 & 83.1 \\ 
Symm-R18~\cite{dudar2019use}$^*$ & 3.8M & 11.6G & 82.3 & 46.1 & 69.2 & 93.9 & 93.8 & 93.4 & 91.9 & 79.3 & 89.7 & {\ul 94.6} & 90.4 & 93.8 & 81.4 & 71.9 & 81.2 \\ \midrule \rowcolor[HTML]{EFEFEF} 
\convabb-R18 & 3.9M & 6.2G & 88.9 & \textbf{75.8} & \textbf{88.6} & 97.2 & \textbf{97.1} & {\ul 97.2} & 95.7 & \textbf{94.5} & \textbf{95.7} & \textbf{95.6} & \textbf{95.2} & \textbf{95.6} & {\ul 85.9} & \textbf{84.9} & \textbf{85.9} \\
\bottomrule
\end{tabular}
}
\caption{\textbf{Main Evaluation Results.} We measure the accuracy (\%) of each baseline on the CIFAR-10~\cite{krizhevsky2009learning}, NWPU-10~\cite{su2019object}, MTARSI~\cite{wu2020benchmark}, NCT-CRC~\cite{kather2019predicting}, and PatchCamelyon~\cite{bejnordi2017diagnostic} datasets. We train each model without rotation and evaluate the test set without rotation (Orig.), with rotations (Rot.), and with reflection (Ref.). We report the average accuracy of the rotated test set. We report the \#parameters and NCT-CRC training GPU memory usage for each method as well. We highlight the best performance with bold and the second best with underline. Our method is shaded in gray. $^*$ indicates the model is re-implemented by us (\cref{sec:model_config}).}
\label{tab:main}
\vspace{-4mm}
\end{table*}

\noindent\textbf{Kernel Initialization.}
We modify the initialization of our \convabb kernel for training stability. Following conventional Kaiming Normal initialization~\cite{he2015delving}, we recompute the sampling boundary according to the actual parameter size $(C, n)$. By default, we set $n=\frac{k+1}{2}$. Meanwhile, we initialize the standard deviations $\Theta_\sigma$ so that the FWHM equals ring width, \ie, $\frac{\Delta d}{2.355}$.
Finally, we clip $\Theta_\sigma$ to the range $[1\times10^{-2}, 2n]$ to ensure a meaningful range of Gaussian functions.

\noindent\textbf{Parameter Efficiency.} Compared to a standard CNN kernel of size $(k,k)$, our \convabb kernel reduces the number of parameters exponentially from $O(C\cdot k^2)$ to $O(C\cdot n)$ and thus compresses the model size. Experiments demonstrate that our method can maintain SoTA performance on various datasets despite fewer parameters. 
This approach also enables the efficient parameterization of large kernels with a large effective receptive field~\cite{Luo2017-sy}.

\subsection{Computational Efficiency}
\label{sec:method_comp}

We note that our \convname kernel would bring extra computational cost if using the standard convolution operation since we need to compute the full $K_{\convabb}$ first, just like other native radial symmetric methods~\cite{fuhl2021rotated,dudar2019use,yeh2016stable}. However, our unique design of shared Gaussian rings across each channel allows us to decompose the full convolution operation into two separate filter-wise operations with higher computational efficiency (\cref{fig:build}(b)).
Namely, we convolve the input first with each discretized Gaussian ring in $M_G$, and then apply the $\Theta_w$ weighted $1\times1$ channel convolution.

\convname can be viewed as the $\Theta_w=[w_1, w_2, \dots, w_n]$ weighted sum of $n$ Gaussian mixture rings $G_i$ (\cref{sec:gmr}). Thus, we can first convolve the input with each Gaussian mixture ring and then compute the weighted sum using a 1D grouped convolution parameterized by $\Theta_w$ (\cref{fig:build}(b)). This operation is strictly equivalent to convolving the input with the complete $K_{\convabb}$ due to the associative property of convolution, where the matrix multiplication in \cref{eq:discrete_kernel} can be viewed as a convolution with a kernel of shape 1-by-1.

The computational complexity of our \convname is $O(HWn(k^2+C_{in}C_{out}))$ (see \cref{sec:complexity}), while the complexity of standard convolution is $O(HWk^2C_{in}C_{out})$, where $H$ and $W$ are the input image height and width, respectively. Considering that both $n$ and $k$ are usually small integers, the computational complexity is greatly reduced as long as $n < C_{in}C_{out}$ and $n < k^2$. Because $n$ is defined to be strictly smaller than $k$, the efficient implementation of \convname is faster in the actual computation (\cref{fig:teaser}(b), \cref{sec:speed}).

\begin{figure*}[t]
    \centering
    \includegraphics[width=1.0\textwidth]{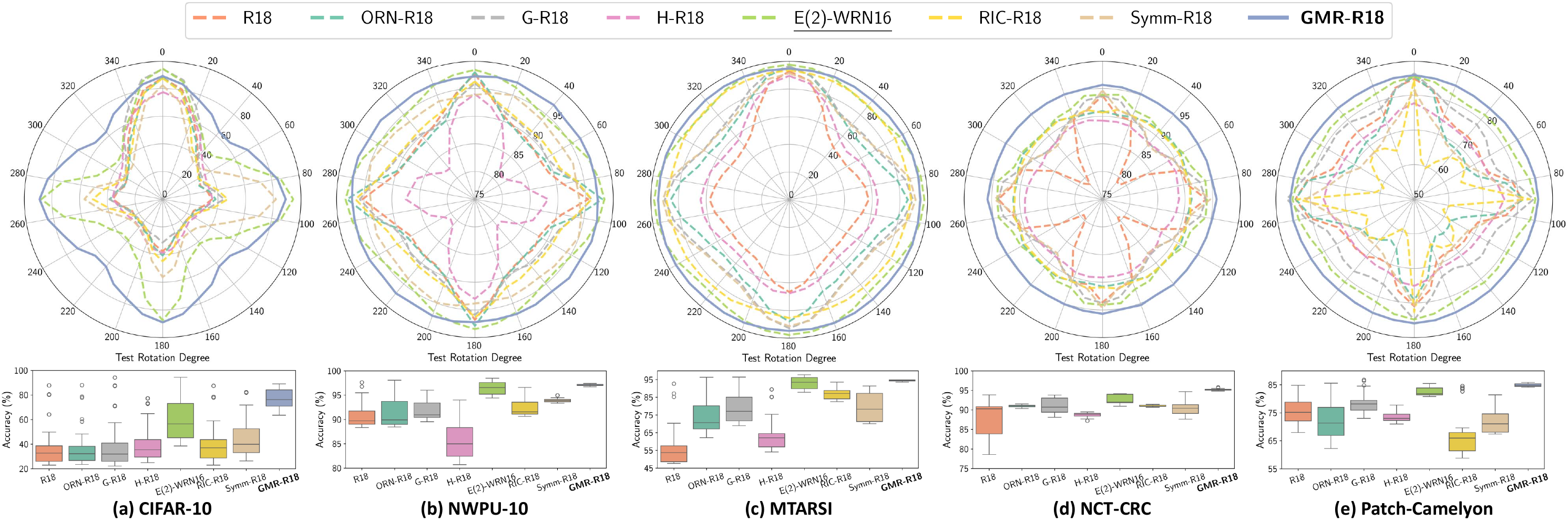}
    \caption{{\bf Classification Accuracy Across Rotation Angles.} We visualize the classification accuracy (\%) per test rotation angles and the box plot of per-angle accuracy for (a) CIFAR-10~\cite{krizhevsky2009learning}, (b) NWPU-10~\cite{su2019object}, (c) MTARSI~\cite{wu2020benchmark}, (d) NCT-CRC~\cite{kather2019predicting}, and (e) Patch-Camelyon~\cite{bejnordi2017diagnostic}. We scale the radial axis to better visualize differences in performance. We omit the E(2) baseline with only $0^{th}$-order filters as its performance is much lower than the E(2)-WRN16. Symm-R18 serves as the naive radial symmetry convolution baseline.}
    \label{fig:radar3}
    \vspace{-3mm}
\end{figure*}  
\mysection{Experiments and Results}

We validate our method's effectiveness on eight different datasets with five real-world applications where the image has no specific orientation. We compare with multiple rotation-equivariant baselines trained without rotation and test them on rotated and reflected test sets as done in prior work~\cite{weiler2019general,mo2022ric,dudar2019use,fuhl2021rotated,kim2020cycnn}. A robust equivariant model is expected to generalize well on both test sets.

\mysubsection{Experimental Setup}
\label{sec:implement}

\noindent\textbf{Datasets.} 
We evaluate the capability of our approach to capture equivariance on four orientation-dependent and five orientation-independent imaging datasets.
First, we experiment on the orientation-dependent, real-world \textbf{CIFAR-10}~\cite{krizhevsky2009learning} dataset that includes  50K training images and 10K test images from 10 different classes.
Next,  we benchmark performance on four orientation-independent imaging datasets. We evaluate two remote sensing datasets: \textbf{NWPU-10}~\cite{su2019object}, which contains 10 different classes of objects, and \textbf{MTARSI}~\cite{wu2020benchmark}, which contains 20 aircraft types cropped from remote sensing pictures. Following Mo~\etal\cite{mo2022ric}, we randomly sample 100 images per class from NWPU-10 and 200 images per class from MTARSI to form the training sets. We further evaluate our method on two histopathological image datasets: a colorectal cancer dataset \textbf{NCT-CRC}~\cite{kather2019predicting} with 100,000 training images for 9 different classes and 7,180 test images, and a lymph node dataset \textbf{Patch-Camelyon}~\cite{bejnordi2017diagnostic} with 262,144 training and 32,768 test images with binary labels for malignancy. 
As an additional evaluation, we use \textbf{ImageNet-1k}~\cite{li2009imagenet} with over 1.2M training images and its validation set to demonstrate the capability of the model with more complex training data in \cref{sec:imagenet}. 
For all datasets, training data is not rotated, while the test data are rotated in 10-degree increments to form a rotated test set, resulting in 36 times the number of original test images, following the standard evaluation procedure~\cite{zhou2017oriented,cohen2016group,worrall2017harmonic,weiler2019general,mo2022ric,dudar2019use}.
We create reflected versions of these 2D datasets by flipping the test images both vertically or horizontally, resulting in twice the number of original test images. 
We also evaluate the capability of \convname on the nuclei segmentation dataset \textbf{CryoNucSeg}~\cite{mahbod2021cryonuseg} in \cref{sec:seg}.
We further extend our method to 3-dimensional convolution and evaluate it on the orientation-dependent  3D \textbf{ModelNet-10} and \textbf{ModelNet-40}~\cite{wu20153d} datasets in \cref{sec:modelnet}.

\noindent\textbf{Evaluation Metrics.}
We evaluate classification performance by assessing accuracy on the following: (1) the original test set; (2) the rotated test set both at each rotation angle and averaged across all angles; and (3) the reflected test set.

\noindent\textbf{\convabb-CNN Model Architecture.} 
We plug our \convname into the classical ResNet18 \cite{he2016deep} as our baseline: \textbf{\convabb-R18}. Following \cref{sec:rules}, we replace the conventional convolutional layers with our \convabb convolutional layers with a kernel width of $9$ for the first two stages and a kernel width of $5$ for the last two stages, and further replace the downsampling blocks with an average pooling layer and a 1-by-1 convolutional layer. 
For CIFAR-10, we remove the first convolution downsampling layer to maintain reasonable feature map dimensions for this dataset's small input size. 
Detailed settings of \convabb-R18 are in \cref{sec:model_config,}. 
We evaluate the effect of different kernel configurations (\cref{sec:ablation}) and the influence of training rotation augmentation (\cref{sec:rot_aug}).

\begin{figure}[!t]
    \centering
    \includegraphics[width=\columnwidth]{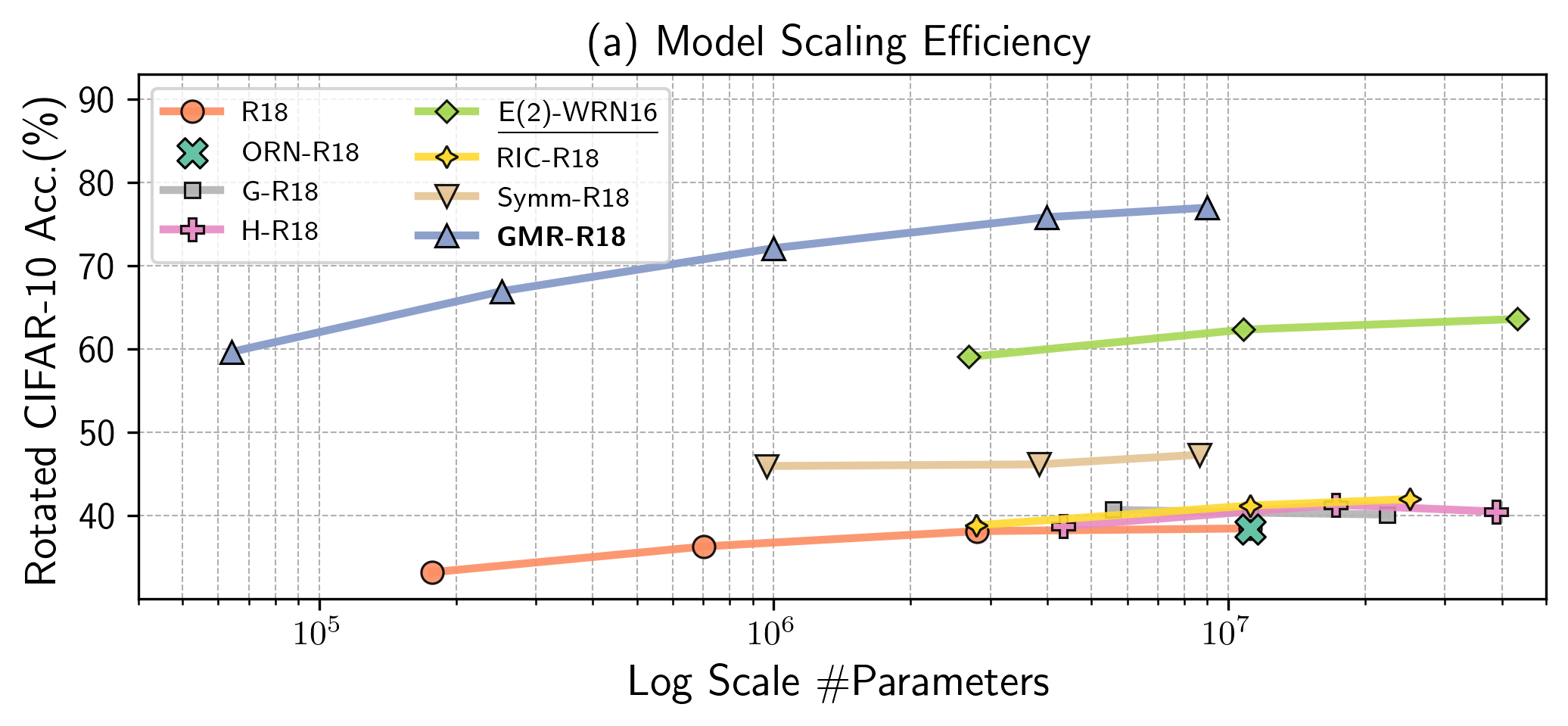}
    \vspace{-7mm}
    \caption{{\bf Model Parameter Efficiency Evaluation}. We plot the averaged accuracy on rotated CIFAR-10 test sets for each model, varying its channel size and number of parameters.}  
    \label{fig:param}
    \vspace{-5mm}
\end{figure}  

\noindent\textbf{Baseline CNNs.~} 
We use conventional \textbf{ResNet18 (R18)}~\cite{he2016deep} as a baseline. We further selected the following SoTA methods with a ResNet~\cite{he2016deep,zagoruyko2016wide}-style architechture: \textbf{ORN-R18}~\cite{zhou2017oriented}, \textbf{G-R18}~\cite{cohen2016group} (G for Group), \textbf{H-R18}~\cite{worrall2017harmonic} (H for Harmonic), \textbf{E(2)-WRN16}~\cite{weiler2019general,cesa2022program} (WRN16 for wide ResNet16~\cite{zagoruyko2016wide}), \textbf{RIC-R18}~\cite{mo2022ric}, and \textbf{Symm-R18}~\cite{dudar2019use}. 
Also, we compared with E(2)-WRN16 using only $0^{th}$-order harmonic filters, a special case of the E(2) network that uses only radially symmetric filters.
We use the corresponding official implementations (except for Symm-R18, which is re-implemented by us) and evaluate under the same settings. We further evaluate all the models with different numbers of parameters by varying the base channel number to illustrate the influence of varying model sizes. We also compare with reported results from the symmetric kernel method \textbf{RAD-ResNet}~\cite{fuhl2021rotated} in \cref{sec:fuhl} due to insufficient re-implementation details.
Details for each model are in \cref{sec:model_config}.

\noindent\textbf{Implementation Details.} We train each model for 250 epochs with the SGD optimizer and cosine annealing scheduler and learning rate of $2\times10^{-2}$ with cross-entropy loss. For NCT-CRC and Patch-Camelyon, we train the model for 50 and 100 epochs, respectively, due to its larger size. We use the AdamW~\cite{loshchilov2017decoupled} optimizer for the H-Net~\cite{worrall2017harmonic} to ensure convergence. All experiments were done with one NVIDIA A5000 GPU. Images from CIFAR-10 are resized to $(32, 32)$, while the images from the remote sensing datasets are resized to $(64, 64)$. 
A batch size of $128$ is used for these three datasets. 
For NCT-CRC, we use an image size of $(224, 224)$ with a batch size of $24$.
For Patch-Camelyon, we use an image size of $(96, 96)$ with a batch size of $64$.
As is standard practice for equivariant feature learning~\cite{zhou2017oriented,cohen2016group,worrall2017harmonic,weiler2019general,mo2022ric,dudar2019use}, no geometric data augmentation is applied during training in order to demonstrate the full capabilities of equivariant learning without introducing confounding effects. The implementation details for ImageNet-1k, ModelNet, and CryoNucSeg is in \cref{sec:add_impl}.

\subsection{CIFAR-10 Results}
\label{sec:cifar_exp}
On CIFAR-10, which contains orientation-dependent real-world images, our method achieves the best performance in both rotated and reflected evaluation with a notable improvement of $13\%$ with less than 34\% of the trainable parameters of competing models (\cref{tab:main}). 
Visualizing test performance across each rotation angle, our model achieves the overall best level of performance and smallest variance across all rotations (\cref{fig:radar3}(a)), and it generally experiences much less performance drop when ``intermediate'' rotation is applied, \eg, 
45{\textdegree}, 135{\textdegree}, 225{\textdegree}, 315{\textdegree}.
Part of the performance loss in these ``intermediate'' angles is caused by border zero padding in rotations. 
Different from other methods that ``compute'' to achieve rotational equivariance, our model is strictly rotational and reflection equivariant by design. Our Gaussian ring smoothing and the circle kernel design help the model generalize on the ``intermediate'' rotations. 
For the reflection transforms, all models show a more robust performance overall since reflection is more realistic in  CIFAR-10.

E(2)-WRN16~\cite{weiler2019general}, with both rotation and reflection equivariance, shows a similar periodic generalization ability to rotations as our method~(\cref{fig:radar3}(a)) but with worse consistency.
E(2) with $0^{th}$-order filters and Symm-R18~\cite{dudar2019use} utilizing na\"{i}ve radial symmetric kernels, perform worse in the original test set and fail to be consistent in the rotation evaluation. This is mainly due to kernel discretization error.

Even with the symmetric constraints, benefiting from the Gaussian mixture ring, our model shows comparable performance on the original test set with the standard ResNet18~\cite{he2016deep}. In contrast, Symm-R18~\cite{dudar2019use} with symmetric constraints performs worse on the original test set.

\begin{table}[!t]
\centering
\setlength{\tabcolsep}{4pt}
\resizebox{0.9\linewidth}{!}
{
\begin{tabular}{lccccc}
\toprule
\multicolumn{1}{c}{\textbf{Methods}} & \multicolumn{1}{c}{\textbf{\#Param}} & \multicolumn{1}{c}{\textbf{Memo.}} & \textbf{~~Orig.~~} & \textbf{~~Rot.~~} & \textbf{~~Ref.~~}\\ \midrule
ResNet50~\cite{he2016deep} & 25.5M & 19.3G & \textbf{74.20} & 28.41 & 57.24 \\
E(2)-ResNet50~\cite{weiler2019general} & 26.1M & 23.7G & 73.45 & 42.39 & 70.49 \\ 
Symm-ResNet50~\cite{dudar2019use} & 18.0M & 19.4G & 63.50 & 53.81 & 62.94 \\ \midrule \rowcolor[HTML]{EFEFEF} 
\convabb-ResNet50 & 18.1M & 18.9G & 73.09 & \textbf{69.68} & \textbf{73.09} \\ \bottomrule
\end{tabular}
}
\caption{\textbf{ImageNet-1k~\cite{li2009imagenet} Results.} We compare the Top-1 accuracy performance (\%) on ImageNet-1k~\cite{li2009imagenet}. We highlight the best performance with bold. Our model is shaded in gray.}
\label{tab:imagenet}
\vspace{-4mm}
\end{table}

\subsection{Remote Sensing Results} 
On the real-world NWPU-10~\cite{su2019object} and MTARSI~\cite{wu2020benchmark} remote sensing datasets, which have no explicit orientation, our model outperforms all baselines on the rotated test set and performs well on the reflected test set (\cref{tab:main}). Our model shows competitive performance on the original test set compared with E(2)-WRN16~\cite{weiler2019general} with only 34\% of the model size. 
We visualize the per-angle accuracy on these datasets in \cref{fig:radar3}(b, c). All methods show a higher average accuracy and a more consistent per-angle performance compared to CIFAR-10 results due to the orientation-independent nature of the remote sensing data. Our model again shows the best consistency and overall best performance. The E(2)-WRN16 baseline failed to maintain rotation equivariance on the intermediate angles despite marginally higher accuracy on the original, unrotated test images. Trivial radial symmetric methods like E(2) with $0^{th}$-order filters and Symm-R18 either perform poorly or fail to be equivariant on both datasets.

\subsection{Histopathology Results} 
On NCT-CRC~\cite{kather2019predicting} and Patch-Camelyon~\cite{bejnordi2017diagnostic}, two large datasets with larger images and no explicit orientation, our method shows consistently competitive performance across all three test settings with up to a $\sim$3\% gap (\cref{tab:main}). 
Per-angle performance on NCT-CRC~(\cref{fig:radar3}(d)) and on Patch-Camelyon (\cref{fig:radar3}(e)) both demonstrate our method's consistent performance across all rotations, highlighting its strong performance on orientation-independent images.

\mysubsection{ImageNet-1k Results}
\label{sec:imagenet}
As one of the most commonly used benchmarks, the evaluation on ImageNet-1k~\cite{li2009imagenet} helps illustrate the capability of each model in more complex real-world settings (\cref{tab:imagenet}). We switch to a ResNet-50 backbone for better performance. Our model is only 1.1\% from the best baseline on the original test set while performing much better in the rotated test set. While E(2)-ResNet50 has a larger model size and demands more GPU memory during training, it is 27\% lower than ours in the rotated test set. Symm-ResNet50 baseline with smaller na\"{i}ve radial symmetry kernel struggles to match the performance of other models. 
This highlights the effectiveness of our smoothed kernel design, enabling a larger receptive field, which is the key to adapting radial symmetric kernel to complex tasks and alleviating the information loss issue. We provide a more comprehensive discussion in \cref{sec:tradeoff}.

\begin{table}[!t]
\centering

% \vspace{-2mm}
\resizebox{\linewidth}{!}
{
\begin{tabular}{cccccccc}
\toprule
\multirow{2}{*}{~\textbf{CK}~} & \multirow{2}{*}{~\textbf{w/o GMR}~} & \multicolumn{3}{c}{\textbf{GMR} $\Theta_\sigma$} & \multicolumn{3}{c}{\textbf{CIFAR-10}~\cite{krizhevsky2009learning}} \\ \cmidrule(l){3-5} \cmidrule(l){6-8}
& & ~Fixed~ & ~Layer~ & ~Ch.~ & Orig. & Rot. & \#Param. \\ \midrule
 & \checkmark &  &  &  & 87.85 & 66.21 & 5.2M \\
 & & \checkmark &  &  & 88.19 & 67.18 & 5.2M \\
 & &  & \checkmark &  & 88.59 & 70.98 & 5.2M \\
 & &  &  & \checkmark & 87.18 & 70.77 & 10.2M \\ \midrule
\checkmark & \checkmark &  &  &  & 87.42 & 72.67 & 3.9M \\
\checkmark & & \checkmark &  &  & 86.57 & 70.89 & 3.9M \\ \rowcolor[HTML]{EFEFEF} 
\checkmark & & & \checkmark &  & \textbf{88.96} & \textbf{75.82} & 3.9M \\
\checkmark & &  &  & \checkmark & 86.36 & 73.31 & 8.0M \\ \bottomrule
\end{tabular}
}
\caption{\textbf{Kernel Design Ablation.} Our full model is highlighted in gray. CK, Circular Kernel; GMR $\Theta_\sigma$, Gaussian Mix Ring; Layer, layer-wise; Ch., channel-wise. }
\label{tab:arch}
% }
% \vspace{-0.5\baselineskip}
\vspace{-5mm}
\end{table}

\begin{table*}[!t]
\centering

\setlength{\tabcolsep}{8pt}
\resizebox{0.9\textwidth}{!}
{
\begin{tabular}{lcccc|ccc|ccc|ccc|ccc}
\toprule
\multicolumn{1}{c}{\multirow{2}{*}{\textbf{Models}}} & \multicolumn{1}{c}{\multirow{2}{*}{\textbf{Rot. Aug}}} & \multicolumn{3}{c|}{\textbf{CIFAR-10}~\cite{krizhevsky2009learning}} & \multicolumn{3}{c|}{\textbf{NWPU-10}~\cite{su2019object}} & \multicolumn{3}{c|}{\textbf{MTARSI}~\cite{wu2020benchmark}} & \multicolumn{3}{c|}{\textbf{NCT-CRC}~\cite{kather2019predicting}} & \multicolumn{3}{c}{\textbf{Patch-Camelyon}~\cite{bejnordi2017diagnostic}} \\ \cmidrule(l){3-17} 
\multicolumn{1}{c}{} & \multicolumn{1}{c}{} & Orig. & Rot. & Ref. & Orig. & Rot. & Ref. & Orig. & Rot. & Ref. & Orig. & Rot. & Ref. & Orig. & Rot. & Ref. \\ \midrule
R18~\cite{he2016deep} &  & {\ul 87.7} & 38.4 & 64.2 & \textbf{97.6} & 90.9 & 95.3 & 92.7 & 57.2 & 65.4 & 93.7 & 87.3 & 92.9 & 84.8 & 75.3 & 82.6\\
R18~\cite{he2016deep} & \checkmark & 87.5 & {\ul 86.2} & 85.9 & {\ul 97.3} & \textbf{97.3} & {\ul 97.1} & 94.6 & 94.0 & 93.5 & 92.0 & 92.2 & 91.6 & 85.8 & 84.2 & 84.3 \\ \midrule \rowcolor[HTML]{EFEFEF}
\convabb-R18 &  & \textbf{88.9} & 75.8 & \textbf{88.6} & 97.2 & {\ul 97.1} & \textbf{97.2} & \textbf{95.7} & {\ul 94.5} & \textbf{95.7} & \textbf{95.6} & \textbf{95.2} & \textbf{95.6} & {\ul 85.9} & {\ul 84.9} & {\ul 85.9} \\ \rowcolor[HTML]{EFEFEF}
\convabb-R18 & \checkmark & 87.4 & \textbf{86.7} & {\ul 87.4} & 96.2 & 96.3 & 96.2 & {\ul 95.0} & \textbf{94.6} & {\ul 95.0} & {\ul 95.2} & {\ul 94.5} & {\ul 95.2} & \textbf{86.6} & \textbf{85.4 }& \textbf{86.6} \\\bottomrule
\end{tabular}
}
\caption{\textbf{Rotation Training Augmentation Evaluation Results.} We compare the performance of our method against ResNet~\cite{he2016deep} trained with random rotations $[-180^\circ,180^\circ]$  on each dataset: CIFAR-10~\cite{krizhevsky2009learning}, NWPU-10~\cite{su2019object}, MTARSI~\cite{wu2020benchmark}, NCT-CRC-100k~\cite{kather2019predicting}, and PatchCamelyon~\cite{bejnordi2017diagnostic}. We evaluate the test set without rotation (Orig.), with rotations (Rot.), and with reflection (Ref.). We report the average accuracy (\%) of each test set. We highlight the best performance in bold and the second in underline. Our method is shaded in gray.}
\label{tab:aug}
\vspace{-4mm}
\end{table*}

\mysubsection{Evaluating Parameter Efficiency}
\label{sec:param_exp}

To further evaluate the parameter efficiency of our model, we train the same \convabb-R18 model and all the baselines on CIFAR-10 with different numbers of base channels (detailed in \cref{sec:model_config}) and evaluate their performance on the rotated test set. The parameter efficiency of each model is shown in \cref{fig:param}. 
We note that our model maintains the same level of performance even when reducing the number of base channels to $8$. Furthermore, our model consistently achieves the best performance over different parameter levels. Compared to a large model like E(2)-WRN16~\cite{weiler2019general}, our model achieves the same level of performance with one-tenth parameters.

\mysubsection{Evaluating Computational Efficiency}
We first evaluate the convolution-only speed in \cref{fig:teaser}(b) and \cref{sec:speed}, where our method is the fastest one under 4 out of 5 kernel widths, realizing a 10 - $100\times$ speed-up compared with Group- and E(2)-Conv. 
In the training setting, the harmonic computation (H-R18 and E(2)-WRN16) and group-based (G-R18 and E(2)-R18) methods require more than three times the GPU memory during training. 
E(2)-WRN16~\cite{weiler2019general} is two times slower than our method ($\sim$450 FPS vs.~$\sim$860 FPS) during training due to its additional computation. 
Meanwhile, the RIC-R18 baselines that require less GPU memory perform sub-optimally. This demonstrates the potential of our model to be applied to edge computing devices where computational resources may be limited.

\begin{table}[!t]
\centering

\setlength{\tabcolsep}{4pt}
\resizebox{0.78\linewidth}{!}
{
\begin{tabular}{lccc|cc}
\toprule

\multicolumn{1}{c}{\multirow{2}{*}{\textbf{Models}}} & \multicolumn{1}{c}{\multirow{2}{*}{\textbf{\#Param}}} & \multicolumn{2}{c|}{\textbf{ModelNet-10}} & \multicolumn{2}{c}{\textbf{ModelNet-40}} \\ \cmidrule(l){3-6} 
\multicolumn{1}{c}{} & \multicolumn{1}{c}{} & Orig. & Rot. & Orig. & Rot. \\ \midrule

R3D~\cite{tran2018closer} & 2.02M & \textbf{92.3} & 42.0 & \textbf{91.4} & 33.5 \\
SE(3)-CNN~\cite{fuhl2021rotated} & 0.23M & 87.1 & 55.4 & 78.9 & 50.5 \\
Symm. R3D~\cite{fuhl2021rotated} & 0.25M & 88.4 & {\ul 55.6} & 85.1 & {\ul 52.3} \\ \midrule \rowcolor[HTML]{EFEFEF} 
\convabb-R3D & 0.17M & {\ul 90.5} & \textbf{65.4} & {\ul 88.0} & \textbf{61.9} \\ \bottomrule
\end{tabular}
}
\caption{\textbf{3D ModelNet~\cite{wu20153d} Results.} We compare the 3D object classification performance on ModelNet-10 and -40~\cite{wu20153d}. We use the provided SE(3)-CNN~\cite{cesa2021program}. All models use the same architecture as SE(3)-CNN. We highlight the best performance with bold and the second-best with an underline. Our model is shaded in gray.}
\label{tab:modelnet}
\vspace{-3mm}
\end{table}

\mysubsection{Evaluating Rotational Data Augmentation}
\label{sec:rot_aug}

We compare our method with conventional R18~\cite{he2016deep} with random training rotation augmentation $[-180^\circ,180^\circ]$  (\cref{tab:aug}). 
Training augmentation improves R18 performance under rotated and reflected test sets; however, \convabb-R18 generally outperforms the conventional CNN with data augmentation in orientation-independent datasets, even without augmentation. Meanwhile, we note that random rotation may harm model performance on the original test set, highlighting the necessity of native equivariance.
Even in CIFAR-10, where images have orientation, our method performs better with much fewer resources than the baseline with the same augmentation. 
We visually compare R18 and \convabb-R18 feature spaces using t-SNE~\cite{van2008visualizing} for the NCT-CRC dataset in \cref{sec:eq_feat}. We also evaluate the models' performance with different levels of augmentations in \cref{sec:partial}.

\mysubsection{Ablation Study}
\label{sec:ablation}

We evaluate different \convname configurations on the rotated CIFAR-10 dataset (\cref{tab:arch,tab:ablation1}) with detailed analysis provided in the Supplement (\cref{sec:kernel_design,sec:kernel_config}). We first evaluate the influence of using different designs proposed in \cref{tab:arch}. We note that introducing the circular kernel design consistently improves the performance on the rotated test set by $\sim$2\%. Due to discrete artifacts, a standard symmetric kernel with no smoothing underperforms compared to our method. Using a fixed Gaussian smoothed kernel with fixed FWHM $\Theta_\sigma$ improves the performance but is still suboptimal. However, making it a layer-wise trainable parameter immediately boosts the performance by adding just a few hundred trainable parameters. In contrast, making $\Theta_\sigma$  channel-wise trainable, \ie, $\Theta_\sigma\in\mathbb{R}^{(C, n)}$, not only doubles the number of parameters but harms the performance. We speculate that over-parametrization hindered model convergence.

\begin{table}[!t]
\centering

\setlength{\tabcolsep}{8pt}
\resizebox{\columnwidth}{!}
{
\begin{tabular}{lccc|cc|cc}
\toprule
\multicolumn{1}{c}{\multirow{2}{*}{\textbf{Models}}} & \multicolumn{1}{c}{\multirow{2}{*}{\textbf{\#Param}}} & \multicolumn{2}{c|}{\textbf{AUC}} & \multicolumn{2}{c|}{\textbf{IoU}} & \multicolumn{2}{c}{\textbf{DICE}} \\ \cmidrule(l){3-8} 
\multicolumn{1}{c}{} & \multicolumn{1}{c}{} & Orig. & Rot. & Orig. & Rot. & Orig. & Rot. \\ \midrule
U-Net~\cite{ronneberger2015u} & 0.48M  & 95.9 & 92.3 & \textbf{67.5} & 63.9 & \textbf{80.3} & 77.7 \\
E(2) U-Net~\cite{cesa2021program} & 0.37M & \textbf{96.6} & {\ul 94.3} & 66.0 & {\ul 65.4} & 79.1 & {\ul 78.7} \\ 
Symm. U-Net~\cite{dudar2019use} & 0.11M & 95.6 & 91.3 & 60.9 & 60.4 & 75.1 & 74.8\\ \midrule \rowcolor[HTML]{EFEFEF} 
\convabb U-Net & 0.18M & {\ul 96.0} & \textbf{94.8} & {\ul 66.6} & \textbf{65.9} & {\ul 79.8} & \textbf{79.2} \\
\bottomrule
\end{tabular}

}
\caption{\textbf{Nuclei Segmentation Results.} We report the segmentation results on CryoNuSeg~\cite{mahbod2021cryonuseg} dataset with or without $[-180^\circ, 180^\circ]$ degree rotation. We report AUC, IoU, and DICE as our metrics. We highlight the best performance with bold and the second best with unan derline. Our method is shaded in gray.}
\label{tab:seg}
\vspace{-5mm}
\end{table}

\mysubsection{Scaling into 3-Dimensional Data}
\label{sec:modelnet}
We evaluate the model's performance on the 3D object recognition dataset, ModelNet-10 and -40~\cite{wu20153d} (\cref{tab:modelnet}). We use a similar R3D network design~\cite{tran2018closer} and replace its convolutional layers with our \convname with a similar size of the SE(3)-CNN~\cite{cesa2021program} baseline. We rotated the test set every 30 degrees in each axis, forming a test set of 36 times more test data. Our model performs the second best in the original test set and beats all the baselines in the rotated test set, demonstrating the scaling capability of \convname in higher-dimensional data. The advantage of our method in 2D space still remains in the 3D case. The smoothed 3D kernel will help the convolution to generalize better under rotation and reflection transformations. It can also be extended to 2+1D video analysis domain, where we can use a simple combination of 2D \convname and conventional $1\times1$ convolutional layer.

\mysubsection{Nuclei Segmentation Results}
\label{sec:seg}

To evaluate \convname's capability in complex tasks beyond classification, we evaluate segmentation performance on the nuclei CryoNucSeg~\cite{mahbod2021cryonuseg} dataset using the U-Net~\cite{ronneberger2015u} design (\cref{tab:seg}). 
We replace the convolutional layer in the regular U-Net~\cite{ronneberger2015u} with different equivariant convolutional layers as our baselines (details in \cref{sec:impl_cryonucseg}). 
We rotate the test set every 10 degrees to form the rotated test set. 
Our model is the second best in the original test set and performs the best in the rotated test set. It also outperforms the E(2) baseline in 5 out of 6 metrics. This indicates \convname can be applied to segmentation tasks and provides further evidence that radial symmetric \convname can match the performance of regular convolutional layers and improve robustness to rotations. The information loss issue can be mitigated via a simple design like skip connections in the U-Net~\cite{ronneberger2015u}.

\mysection{Discussion and Conclusion}
\label{sec:conclusion}

We propose \convname, a novel radial symmetric rotation and reflection equivariant convolutional kernel, smoothed via a weighted mix of Gaussian functions. Our novel design and efficient implementation enable us to simultaneously achieve consistent equivariance across geometric transformations, high model and computational efficiency, and mitigation of information loss, as shown by our SoTA performance on multiple real-world datasets and tasks. \convname is capable of various complex tasks, achieving an impressive performance even with the radial symmetric constraint. 
Our proposed \convname can be easily integrated into existing CNN architectures, bringing the most consistent equivariance capability and higher efficiency.

\noindent\textbf{Limitations and Future Work.} 
While our \convabb representation achieves nearly the same or even better best-case performance in the original test domain compared with standard convolutional kernels, the symmetry constraint can still limit the model's representation ability and, therefore, impact the general performance.
We plan to enhance model expressivity by introducing a module that can embed local orientation information while maintaining equivariance.

\noindent\textbf{Acknowledgements.}
No funding was received to conduct this study. The authors have no relevant financial or non-financial interests to disclose.

{
    \small
    \bibliographystyle{ieeenat_fullname}
    \bibliography{main}
}
\clearpage
\clearpage
\setcounter{page}{1}
\setcounter{figure}{1}
\setcounter{table}{1}
\setcounter{section}{0}

\maketitlesupplementary

\renewcommand{\thesection}{S\arabic{section}}
\renewcommand{\thefigure}{S\arabic{figure}}
\renewcommand{\thetable}{S\arabic{table}}

\section{Detailed Model Configurations}
\label{sec:model_config}

This section provides detailed model configurations for our methods and each baseline.

\subsection{Standard CNNs}

\paragraph{ResNet~\cite{he2016deep}} We use the official PyTorch ResNet~\cite{he2016deep} model implementation. As mentioned in \cref{sec:implement}, we re-configure the first convolutional layer to have a stride of 1 and remove the first max pooling layer except for the experiment on NCT-CRC~\cite{kather2019predicting} and Patch-Camelyon~\cite{bejnordi2017diagnostic} datasets. This helps maintain the spatial resolution of small input images. In \cref{sec:param_exp}, we experimented with using various base channel numbers, \ie, $[8, 16, 32, 64]$, to demonstrate the influence of different parameters.
We experiment with both ResNet-18 (R18), and ResNet-50 configurations. We have also experimented with the ResNet-34 model in \cref{sec:fuhl}.

\paragraph{ConvNeXt~\cite{liu2022convnet}} We use the official implementation of the PyTorch ConvNeXt~\cite{liu2022convnet} model in \cref{sec:arch}. We also replace the first convolutional layer with a convolutional layer with a stride of 1. Since our goal is to demonstrate the feasibility of our GMR-Conv in different SoTA CNN architectures, we focus our experiment on the small version of the ConvNeXt~\cite{liu2022convnet} model.

\paragraph{DenseNet~\cite{huang2017densely}~} We use the official implementation of the PyTorch DenseNet~\cite{huang2017densely} model in \cref{sec:arch}. We replace the first convolutional layer with a convolutional layer with a stride of 1. We experiment with both DenseNet-121 and DenseNet-161 configurations.

\subsection{Equivariant CNNs}

\paragraph{ORN Model~\cite{zhou2017oriented}} We use the official implementation of the ORN model\footnote{\url{https://github.com/ZhouYanzhao/ORN}}~\cite{zhou2017oriented}. ORN-R18 is configured similarly to the base R18 model.

\paragraph{Group Model~\cite{cohen2016group}} We use the official implementation of the Group Network\footnote{\url{https://github.com/adambielski/GrouPy}}~\cite{cohen2016group}. For the G-R18 model, we use the official implementation from the released code. In \cref{sec:param_exp}, we experimented with base channel number $[32, 64, 96]$.

\paragraph{Harmonic Model~\cite{worrall2017harmonic}} Since the official implementation of the Harmonic network is implemented with the TensorFlow framework, we use the adapted version on PyTorch for a fair comparison\footnote{\url{https://github.com/debjani-bhowmick/harmonic-net-pytorch}}~\cite{worrall2017harmonic}. We implement the H-R18 model based on the original R18 following the official implementation. In \cref{sec:param_exp}, we experimented with base channel number $[32, 64, 96]$.

\paragraph{E(2) Model~\cite{weiler2019general,cesa2021program}} We use the official implementation of the E(2) Network\footnote{\url{https://github.com/QUVA-Lab/e2cnn}}~\cite{weiler2019general}. For the E(2)-WRN model, we also use the officially implemented version. We chose the model with 8-direction equivalence in the first stage and 4-direction equivalence in the following stages since this can best maintain the rotational equivariance. We also configure the E(2)-WRN model to make it reflection equivariant. In \cref{sec:param_exp}, we experimented with base channel number $[32, 64, 96]$. The E(2) baseline with only $0^{th}$-order filters is obtained by setting the maximum frequency of E(2)-Conv layers to zero. This ensures that all the filters are radially symmetric.

\paragraph{RIC Model~\cite{mo2022ric}} We use the official implementation of the RIC models\footnote{\url{https://github.com/HanlinMo/Rotation-Invariant-Coordinate-Convolutional-Neural-Network}}~\cite{mo2022ric}. For the RIC-R18 model, we used the official implementation. After adapting to our framework in \cref{sec:param_exp}, we experimented with base channel number $[32, 64, 96, 128]$ for RIC-R18.

\paragraph{Symm. Model~\cite{dudar2019use}} Since there is no official implementation provided for the Symm-R18~\cite{dudar2019use}, we implement the model following Dudar~\etal\cite{dudar2019use}. We use a fixed kernel size of 3 and follow the given symmetric pattern. Note that we did not change the convolution with a stride larger than 2 since the original implementation keeps it as well. We experimented with base channel number $[32, 64, 96]$.

\paragraph{Radial Convolution (RAD) Model~\cite{fuhl2021rotated}} Similarly, Fuhl~\etal~\cite{fuhl2021rotated} does not provide an implementation of their method, and as re-implementing it from scratch is out of the scope of this paper, we choose to evaluate our method following their settings in \cref{sec:fuhl}.

\paragraph{\convabb Model} 
\label{sec:r2e2_model}
We replace the conventional convolutional layer in the base CNN model with our \convname.
We further adapt the downsampling layers of these models following \cref{sec:rules}. For our \convabb-CNN, we use the kernel configuration as described in \cref{tab:ablation1}, where the first two stages have a larger kernel size of 9, and the latter two have a smaller kernel size of 5. We use the default ring number for all of our models. In \cref{sec:param_exp}, we experimented with base channel numbers $[8, 16, 32, 64, 96]$, where the largest model has a similar parameter level as the standard R18 model. Additionally, we store $\Theta_\sigma$ in logarithmic form to avoid value underflow.

\section{Additional Experiment Settings}
\label{sec:add_impl}

We provide the experiment settings and implementation details for the \textbf{ImageNet-1k}~\cite{li2009imagenet}, \textbf{ModelNet-10}~\cite{wu20153d}, and \textbf{CryoNucSeg}~\cite{mahbod2021cryonuseg} in this section.

\subsection{Implementation Details for ImageNet-1k}
\label{sec:impl_imagenet}
Instead of using the ResNet-18-based model, we choose to use ResNet50~\cite{he2016deep} as our base model. We replace the convolutional layer in ResNet50 with different equivariant convolutional layers in each baseline model. For the E(2)-ResNet baseline, we choose to use a base channel number of $36$, $N=4$, and kernel size of $3\times3$, resulting in a similar number of parameters as regular ResNet50. We train each model for 100 epochs with a starting learning rate of $1\times10^{-1}$, batch size of 96, and cosine learning rate decay with a warm-up of 2 epoch. All the models are optimized using an SGD optimizer and AMP of BFloat-16 precision on 4 RTX A5000 GPUs. Images are resized to $(224, 224)$.

\subsection{Implementation Details for ModelNet}
\label{sec:impl_modelnet}
We use the officially provided SE(3)-CNN~\cite{cesa2021program} as our baseline model and use a ResNet-3D-9~\cite{tran2018closer} as our base architecture. We use the Symm. Conv~\cite{dudar2019use} version of R3D-9 as another baseline to illustrate the behavior of regular radial symmetric kernels. Similarly, we replace the 3D convolutional layers in ResNet-3D with our \convname as our model. All the models are trained for 100 epochs with a learning rate of $2\times10^{-2}$, a batch size of 8, and cosine learning rate decay. We use the SGD optimizer to update the model during training. All the objects are transformed into voxel form and then resized into $(32, 32, 32)$.

\subsection{Implementation Details for CryoNucSeg}
\label{sec:impl_cryonucseg}
We use a regular U-Net~\cite{ronneberger2015u} with 3 downsampling layers and a basis channel size of 16 as our basic model. We also replace the convolutional kernel within the basic U-Net with our \convname of kernel width $[9, 9, 5, 5]$ for each layer. We train each model for 7,000 epochs with randomly cropped image regions of size $(96, 96)$. A learning rate of $2\times10^{-3}$, a batch size of 8, and AdamW~\cite{loshchilov2017decoupled} optimizer are used for training.

\section{Building a CNN with \convname} 
\label{sec:rules}

To build a CNN with our \convname layers (\convabb-CNN), we follow these principles to maintain equivariance to rotation, reflection, and translation:

\mypar{Build a Fully Convolutional Network (FCN)} 
FCNs~\cite{long2015fully} enable the processing of images regardless of input size. 
\convname can be incorporated into the FCN to ensure equivariance.
An FCN has a backbone with no operation along spatial dimensions other than pooling layers.
Since there is no reshaping operation on the spatial dimension, the equivariance in the feature maps obtained from the previous \convabb layers will be maintained for the classifier.

\mypar{Use Convolution Stride of 1} Considering the central symmetric nature of the \convname, rotational equivariance is only maintained in the sub-region in the convolved patch and extends to global equivariance via kernel sliding. However, using a convolutional stride $s>1$ when the input is rotated may cause the kernel to convolve at a different position and, therefore, result in a different, non-equivariant feature map. 
We suggest using a composition of a pooling layer with size and stride $s$ followed by a 1-by-1 convolutional layer with stride 1, which are both rotation-equivariant. This also helps preserve the rotational equivariance at the model level.

\mypar{Use Large Kernels} Given that the number of parameters increases linearly with the kernel width $k$, it is natural to use a larger convolutional kernel for a larger receptive field and better expressivity. This can also mitigate potential degradation due to the symmetry constraint.

\begin{table*}[!t]
\centering

\resizebox{\linewidth}{!}
{
\begin{tabular}{lcc|cc|cc|cc|cc}
\toprule
\multicolumn{1}{c}{\multirow{2}{*}{\textbf{Conv. Methods}}} & \multicolumn{2}{c|}{$\bm{k = 3}$} & \multicolumn{2}{c|}{$\bm{k = 5}$} & \multicolumn{2}{c|}{$\bm{k = 7}$} & \multicolumn{2}{c|}{$\bm{k = 9}$} & \multicolumn{2}{c}{$\bm{k = 11}$} \\ \cmidrule(l){2-11} 
 & Training (s) & Inference (s) & Training (s) & Inference (s) & Training (s) & Inference (s) & Training (s) & Inference (s) & Training (s) & Inference (s) \\ \midrule
Torch Conv & \textbf{0.974} & {\ul 0.391} & {\ul 1.524} & {\ul 0.727} & 3.579 & 1.418 & 5.148 & {\ul 2.347} & {\ul 6.887} & {\ul 3.493} \\
E(2)-Conv~\cite{weiler2019general} & 20.815 & 10.981 & 57.967 & 28.940 & 122.772 & 67.052 & 204.149 & 114.796 & 313.260 & 175.155 \\
Group-Conv~\cite{cohen2016group} & 23.997 & 10.686 & 67.881 & 29.520 & 138.377 & 62.295 & 229.881 & 104.849 & 345.862 & 162.366 \\ 
Symm.-Conv~\cite{dudar2019use} & 1.058 & 0.412 & 1.467 & 0.738 & {\ul 3.112} & 1.429 & {\ul 4.119} & 2.415 & 7.258 & 3.544 \\\midrule
Na\"{i}ve \convabb-Conv & 1.456 & 0.635 & 1.561 & 0.766 & 3.115 & {\ul 0.817} & 4.858 & 2.440 & 7.162 & 3.558 \\ \rowcolor[HTML]{EFEFEF}
Efficient \convabb-Conv & {\ul 1.194} & \textbf{0.276} & \textbf{1.139} & \textbf{0.270} & \textbf{1.680} & \textbf{0.485} & \textbf{2.542} & \textbf{0.689} & \textbf{2.869} & \textbf{0.927} \\ \bottomrule
\end{tabular}
}
% }
% \vspace{-0.5\baselineskip}
\caption{\textbf{Convolution Speed Evaluation.} We evaluate the speed of different convolution methods during training and inference time, where $k$ is the kernel size. We repeat the convolution operation 1,000 times on a random input and compare the total time to complete the computation in seconds. We highlight the fastest results in bold and the second fastest with underline.}
\label{tab:speed}
\vspace{-5mm}
\end{table*}

\section{Trade-offs Between Preserving Equivariance and Local Orientation Information}
\label{sec:tradeoff}

As claimed previously in \cref{sec:sre}, a radial symmetric kernel is invariant to local Hadamard multiplication, \ie, element-wise matrix multiplication. An intuitive concern raised here is that such a locally invariant property can cause a loss of local orientation-dependent texture information and, therefore, harm the performance.

As discussed in the \cref{sec:intro}, there is a trade-off between preserving equivariance and local orientation information.
The local invariance of the radial symmetric kernel always gives the same output under local element-wise matrix multiplication, regardless of the rotation of the local input, \ie, it cannot provide local orientation-dependent information under Hadamard multiplication to the \emph{channel dimension} of the output. However, the sliding window property of convolution allows the model to \emph{keep} the spatial texture information in the \emph{spatial dimension}, as depicted in the \cref{fig:feat}, where all the orientation-dependent texture details are kept in the output feature map. One simple example will be considering convolving the input with a 2D Dirac delta function kernel $\delta(x, y)$, where $\delta(0, 0)=1$ and $0$ everywhere else. This kernel is undoubtedly radially symmetric. The output will be the same as the input due to the nature of the Dirac delta function, and all the local texture and high-frequency information is kept in the spatial dimension.
\begin{equation}
    (\delta * f)(x, y) = \iint \delta(u, v)f(x-u, y-v)dudv = f(x, y)
\end{equation}
Similarly, the radial symmetric Difference of Gaussians (DoG) filter was used for edge detection, which illustrates its capability to capture local information. 

However, what may trigger the problem is when the model encounters pooling layers. Different from conventional convolution that can encode local texture information into the channel dimension, our method of the radial symmetric convolutional kernel will lose the orientation-dependent information due to the pooling layer, as the pooling layer is downsampling over the spatial dimension, which causes the model to be unaware of the local information within each downsampling window. We acknowledge this concern about the loss of local information and consider this as one of the limitations of our method. This also motivates our future improvements as discussed in \cref{sec:conclusion}. 

However, our experimental results in \cref{tab:main}, \cref{tab:imagenet}, \cref{tab:seg}, and \cref{tab:aug} demonstrate that our \convname is capable of achieving similar performance as conventional convolution in multiple complex and real-world applications, even considering the symmetric kernel constraint.
We sacrifice some of the local orientation-dependent information to gain the desired rotation and reflection equivariance property with much better efficiency in terms of model size, GPU memory, and speed, compared with existing methods~\cite{weiler2018learning,weiler2019general,cohen2016group}. Our design balances equivariance and the capability of handling complex real-world tasks with better efficiency.

Furthermore, the spatial pooling operation contributes to achieving the rotation and reflection invariant output, which is beneficial in tasks such as geospatial or histopathological image classification. The model is expected to produce consistent output regardless of the input's rotation or reflection transformation. Besides, as suggested in \cref{sec:gmr}, the discretization of the radial symmetric kernel will harm the invariant or equivariant property since it is no longer perfectly radial symmetric. However, such imperfect discretization also means our radial kernel can capture some degree of local orientation-dependent information. This may also explain why the model performs well on the original test set without rotation. As shown in the \cref{fig:radar3}(a) and \cref{tab:rot_std}, our method is also not perfectly equivariant to test-time rotation on the dataset like CIFAR-10~\cite{krizhevsky2009learning}, where the orientation is meaningful. We believe the most important thing in our \convname is the trade-off between approximation to a perfect continuous radial symmetric kernel and the model's awareness of the local orientation-dependent information. Our proposed \convname shows impressive balancing on both sides, with a much better efficiency.

Additionally, we believe this issue could be addressed easily in more complex tasks like segmentation or object detection by using a U-Net~\cite{ronneberger2015u} or Feature Pyramid Network~\cite{lin2017feature} architecture, respectively, where the model extracts a feature map from each stage before the spatial pooling layer and predicts based on these feature maps from different scales. By adapting our \convname with these architectures, we can avoid the loss of local texture information since we are using all the spatial information from each stage, which means the information loss due to the pooling layer will not influence the final prediction. We can also combine our equivariant feature extractor with regular prediction heads to fully utilize local orientation-dependent information.

\section{Derivation of Computational Complexity}
\label{sec:complexity}
Here, we derive the computational complexity of the proposed efficient \convname as mentioned in \cref{sec:method_comp}. All the convolution operations are done using the \texttt{im2col} algorithm.
Given a 2D input of size $(B, C_{in}, H, W)$, and a convolutional kernel of shape $(C_{in}, C_{out}, k, k)$, the complexity of na\"{i}ve convolution will be $O(HWk^2C_{in}C_{out})$ since the batch dimension is parallelized.

To simplify this convolution process, we first reshape the input into $(BC_{in}, 1, H, W)$ and convolve it with the discretized rings $G_i, i=1,\dots,n$, whose shape is $(1, n, k, k)$. The output will have a shape of $(BC_{in}, n, H, W)$. This is depthwise convolution with almost no channel dimension operation (the number of rings $n$ is much less than the number of channels in standard convolution kernels), whose computational complexity is $O(HWk^2n)$.

We then reshape the output from previous step into $(B, nC_{in}, H, W)$ and convolve with 1-by-1 $\Theta_w$-weighted Gaussian Ring Mixing kernel $\Theta_w$ with shape $(nC_{in}, C_{out}, 1, 1)$, and the output will have the desired shape $(B, C_{out}, H, W)$. This step has a computational complexity of $O(HWnC_{in}C_{out})$.

This process is mathematically equivalent to convolving with the full $K_{\convabb}$ but has no redundant computation. So the overall final computational complexity will be $O(HWn(k^2+C_{in}C_{out}))$.

\begin{table}[!t]
\centering

\resizebox{\linewidth}{!}
{
\begin{tabular}{lccccc}
\toprule
\multicolumn{1}{c}{\multirow{2}{*}{\textbf{Method}}} & \multicolumn{5}{c}{\textbf{Rotation Test Set Performance w/ Standard Deviation (\%)}} \\ \cmidrule(l){2-6} 
\multicolumn{1}{c}{} & \textbf{CIFAR-10} & \textbf{NWPU-10} & \textbf{MTARSI} & \textbf{NCT-CRC} & \textbf{PCam.} \\ \midrule
R18~\cite{he2016deep} & 38.4$\pm$18.7 & 90.9$\pm$2.7 & 57.2$\pm$12.7 & 87.3$\pm$5.1 & 75.3$\pm$4.9 \\ \midrule
ORN-R18~\cite{zhou2017oriented} & 38.4$\pm$18.0 & 91.4$\pm$3.2 & 87.1$\pm$9.6 & 91.0$\pm${\ul 0.3} & 72.0$\pm$6.9 \\
G-R18~\cite{cohen2016group} & 40.3$\pm$21.3 & 91.8$\pm$2.0 & 79.2$\pm$8.5 & 90.8$\pm$2.0 & 78.7$\pm$3.6 \\
H-R18~\cite{worrall2017harmonic} & 41.2$\pm${\ul 15.3} & 85.9$\pm$3.9 & 74.4$\pm$9.7 & 88.7$\pm$0.5 & 73.5$\pm$1.8 \\
E(2)-WRN16~\cite{weiler2019general} & {\ul 62.3}$\pm$18.2 & {\ul 96.3}$\pm$1.4 & {\ul 93.0}$\pm$3.5 & {\ul92.5}$\pm$1.2 & {\ul 82.6}$\pm${\ul 1.5} \\
RIC-R18~\cite{mo2022ric} & 40.3$\pm$17.4 & 92.4$\pm$1.8 & 63.7$\pm${\ul 3.2} & 91.0$\pm$\textbf{0.2} & 67.0$\pm$7.3 \\
Symm-R18$^*$~\cite{dudar2019use} & 46.1$\pm$16.7 & 93.8$\pm${\ul 0.5} & 79.3$\pm$8.0 & 90.4$\pm$2.0 &  71.9$\pm$4.5 \\ \midrule \rowcolor[HTML]{EFEFEF} 
\convabb-R18 & \textbf{75.8}$\pm$\textbf{8.7} & \textbf{97.1}$\pm$\textbf{0.2} & \textbf{94.5}$\pm$\textbf{0.4} & \textbf{95.2}$\pm$\textbf{0.2} & \textbf{84.9}$\pm$\textbf{0.6} \\ \bottomrule
\end{tabular}
}
\caption{\textbf{Rotation Test Set Performance with Standard Deviation.} We provide the rotation test set accuracy with standard deviation of each method in the form of \texttt{Mean$\pm$STD}. The standard deviation is computed among all 36 different rotations. The lower the standard deviation is, the better the model's ability to capture equivariance. We highlight the best performance with bold and the second best with an underline. $^*$ indicates the model is re-implemented by us (\cref{sec:model_config}).}
\label{tab:rot_std}
\vspace{-5mm}
\end{table}

\section{Convolution Speed Comparison}
\label{sec:speed}

We compare the convolution computation speed of each method in~\cref{tab:speed}. We choose to compare our method with standard PyTorch convolution, E(2)-Conv~\cite{weiler2019general} with the number of basis $N=8$, Group-Conv~\cite{cohen2016group}, and Symm.-Conv~\cite{dudar2019use}. Both E(2)-Conv and Group-Conv are SoTA convolutional layers designed for equivariant learning through extra computation. We compare the speed of a single convolution layer with kernel size $k$ on input with shape $(2, 128, 64, 64)$, and the output channel number is also 128. This is a common input shape that might appear in modern CNN architectures. We repeat the convolution with randomly generated input 1,000 times and report the total time for each method. Additionally, we set \texttt{torch.backends.cudnn.benchmark} to False and \texttt{torch.backends.cudnn.deterministic} to True. This ensures that CUDA always uses the default implicit GEMM (General Matrix Multiplication) with pre-compilation. We warm up the GPU with 100 random general convolutions before the evaluation. We transfer the randomly generated input tensor to the corresponding E(2) space before convolving it with the E(2)-Conv to mimic the behavior of the convolution operation in the model. A similar conversion was done for Group-Conv~\cite{cohen2016group}. The converting operation is not counted in the operation time.

We note that when the kernel size is relatively small ($k=3$), our method shows a speed comparable to that of the conventional convolutional layer. Our \convname is slightly slower at training time since it needs double GPU memory access per operation, one for depthwise Gaussian ring convolution and one for 1-by-1 weighted sum convolution. However, when the kernel size increases, our efficient \convname implementation will be $\sim$3 times faster for both training and inference than the conventional convolution, since the complexity of our method increases linearly with respect to kernel size, as proven in ~\cref{sec:complexity}. In comparison, the conventional convolution's complexity grows at a square rate to the kernel size. In this case, with a larger kernel size, the number of memory accesses is no longer the bottleneck, and our efficient \convname is much faster than na\"{i}ve convolution during both training and inference.

On the other hand, other existing equivariant convolutional layers (E(2)-Conv~\cite{weiler2019general} and Group-Conv~\cite{cohen2016group}) are more than 50 times slower than our \convname when having the same kernel size of $k=11$. This is mainly because these methods need to duplicate their channels or convolution output for each pre-defined rotation basis. For example, E(2)-Conv with $N=8$ will have 8 times more channels, and Group-Conv will also have 4 times more channels and, therefore, 8 or 4 times more computation compared with conventional convolution. This will greatly slow down the model's speed and increase its GPU memory cost (\cref{tab:main}).

\begin{figure*}[t!]
    \centering
    \includegraphics[width=0.95\textwidth]{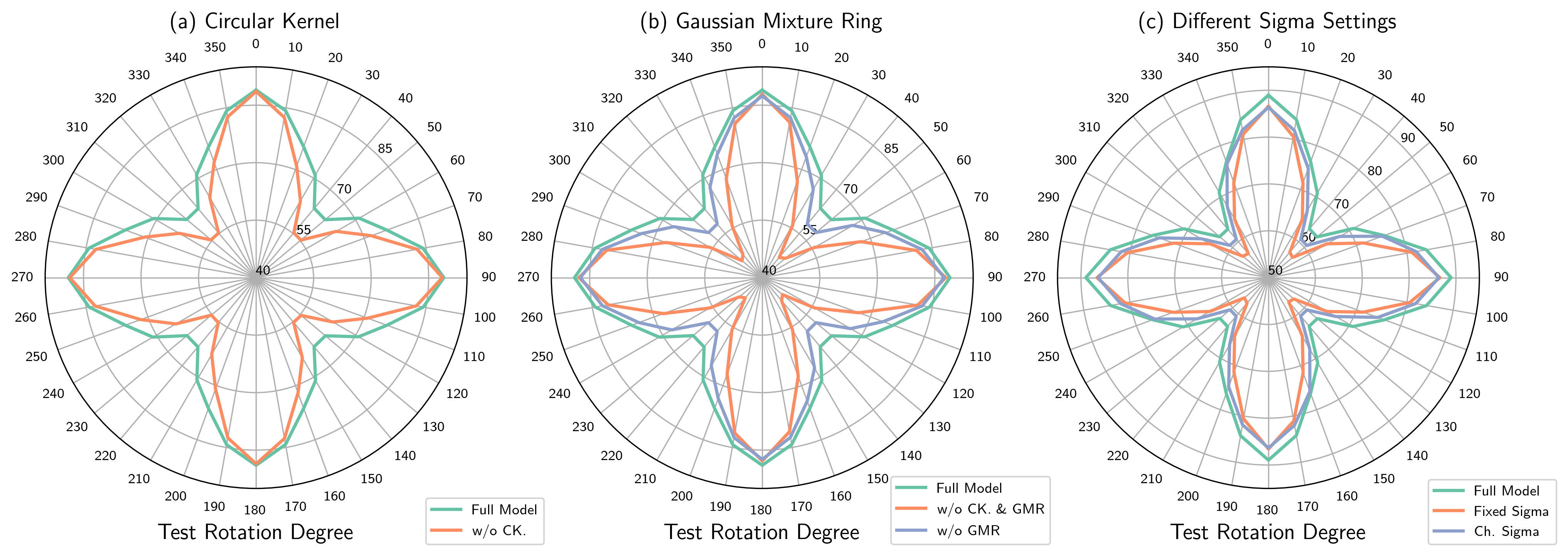}
    \caption{{\bf CIFAR-10~\cite{krizhevsky2009learning} Kernel Design Experiment Across Rotation Degree}. We plot the detailed accuracy curve for each rotation angle in a radar plot for each group of kernel design experiments (\cref{tab:arch}). CK. refers to the circular kernel, GMR refers to the Gaussian Mixture Ring, and Sigma is the parameter $\Theta_\sigma$ that controls each Gaussian function's standard deviation. Our choice of model design shows the best performance.}  
    \label{fig:cifar_kdesign}
    \vspace{-3mm}
\end{figure*}  

\begin{table}[!t]
\centering

\setlength{\tabcolsep}{4pt}
\resizebox{0.75\linewidth}{!}
{
\begin{tabular}{lcc}
\toprule
\multicolumn{1}{c}{\textbf{Methods}} & \textbf{~~4ROT~~} & \textbf{~~ALLROT~~} \\ \midrule
RAD 3-ResNet34~\cite{fuhl2021rotated} & 69.83 & 65.92 \\
RSDW 3-ResNet34~\cite{fuhl2021rotated} & 65.82 & 58.97 \\
RING 5$\times$5-ResNet34~\cite{fuhl2021rotated} & 79.96 & 75.79 \\ \midrule \rowcolor[HTML]{EFEFEF} 
\convabb-ResNet34 & \textbf{90.36} & \textbf{77.86} \\ \bottomrule
\end{tabular}

}
\caption{\textbf{Comparison with Fuhl~\etal\cite{fuhl2021rotated}.} We compare our method with Fuhl~\etal\cite{fuhl2021rotated} on CIFAR-10~\cite{krizhevsky2009learning} following the evaluation metrics used in their work. 4ROT indicates average accuracy over $0^\circ,90^\circ,180^\circ,270^\circ$ rotations; ALLROT indicates average accuracy over a test set with rotation every 20 degrees. All the models follow the ResNet-34~\cite{he2016deep} architecture. We highlight the best performance with bold. Our model is shaded in gray.}
\label{tab:fuhl}
\vspace{-4mm}
\end{table}

\section{Equivariance Analysis}

We further analyze the model's capability to capture rotation equivariance in \cref{tab:rot_std}. We present the standard deviation of each method in the rotation test set, where the test data is rotated every 10 degrees, and compute the standard deviation of the accuracy with respect to the accuracy at each angle. 
Our \convabb method shows the lowest general standard deviation in four out of five datasets (\cref{tab:rot_std}). This demonstrates the impressive capability of our method to robustly maintain rotation invariance across arbitrary rotations. Even compared with methods like E(2)-WRN16~\cite{weiler2019general}, which achieve rotational equivariance through computation, our method shows a more consistent behavior. We omit the analysis for the reflection test set since our method is strictly reflective equivariant due to its symmetric property. To the best of our knowledge, our \convname is one of the best methods for achieving rotation and reflection equivariance.

\section{Comparison with Radial Depth-wise Convolution}
\label{sec:fuhl}
We further compare our model with the method proposed by Fuhl~\etal\cite{fuhl2021rotated}. Since there is no official implementation provided, we follow the evaluation settings as in Fuhl~\etal\cite{fuhl2021rotated} and compare our model with three of the proposed models: Radial Convolutions (RAD) 3, RSDW (Radial Depth-wise Convolution) 3, and the best RING (Rotated Ring Radial Convolution) 5$\times$5 with ResNet34 backbone. Our experimental results show that our model outperforms all baselines proposed by Fuhl~\etal\cite{fuhl2021rotated} (\cref{tab:fuhl}). We demonstrate that the use of a larger \convname kernel and the Gaussian smoothed kernel helps to improve the model's performance in non-interpolated situations, and therefore substantially improves the general performance.

\begin{table}[!t]

\centering
\resizebox{.75\linewidth}{!}
{
\begin{tabular}{cccc}
\specialrule{.1em}{.05em}{.05em}
\multicolumn{2}{c}{\multirow{2}{*}{\textbf{\convname Kernel Config.}}}  & \multicolumn{2}{c}{\textbf{CIFAR-10}~\cite{krizhevsky2009learning}} \\ \cmidrule(l){3-4} 
\multicolumn{2}{c}{}  & ~Rot.~   & ~\#Param.~  \\ \specialrule{.08em}{.05em}{.05em}%\midrule
\multirow{4}{*}{Kernel Size}  & [3, 3, 3, 3]  & 64.86 & 2.6M  \\
  & [5, 5, 5, 5]  & 69.46 & 3.8M  \\
  & [7, 7, 7, 7]  & 74.74 & 5.1M  \\
  & [9, 9, 9, 9]  & \textbf{75.51} & 6.3M  \\ \midrule
\multirow{4}{*}{\begin{tabular}[c]{@{}c@{}}Number of \\ Rings\end{tabular}}   & \cellcolor[HTML]{EFEFEF}[5, 5, 3, 3] & \cellcolor[HTML]{EFEFEF}\textbf{75.82} & \cellcolor[HTML]{EFEFEF}3.9M  \\
  & [5, 5, 4, 4] & 72.98& 5.1M  \\
  & [6, 6, 3, 3] & 73.38& 4.1M  \\
  & [7, 7, 5, 5] & 67.36& 5.4M  \\ \midrule
\multirow{4}{*}{\begin{tabular}[c]{@{}c@{}}Layer-wise \\ Kernel Size\end{tabular}} & \cellcolor[HTML]{EFEFEF}[9, 9, 5, 5]  & \cellcolor[HTML]{EFEFEF}\textbf{75.82} & \cellcolor[HTML]{EFEFEF}3.9M  \\
  & [7, 7, 5, 5] & 74.69& 3.8M  \\
  & [11, 11, 9, 9]   & 74.07& 6.4M  \\
  & [11, 9, 7, 5]& 73.63& 4.2M  \\ \specialrule{.1em}{.05em}{.05em}
\end{tabular}
}
\caption{\textbf{Kernel Shape Ablation.} We present the ablation results with different \convname kernel configurations. The best results are highlighted in bold. Our full model is shaded in gray.}
\label{tab:ablation1}
\vspace{-3mm}
\end{table}
\begin{figure*}[t!]
    \centering
    \includegraphics[width=0.95\textwidth]{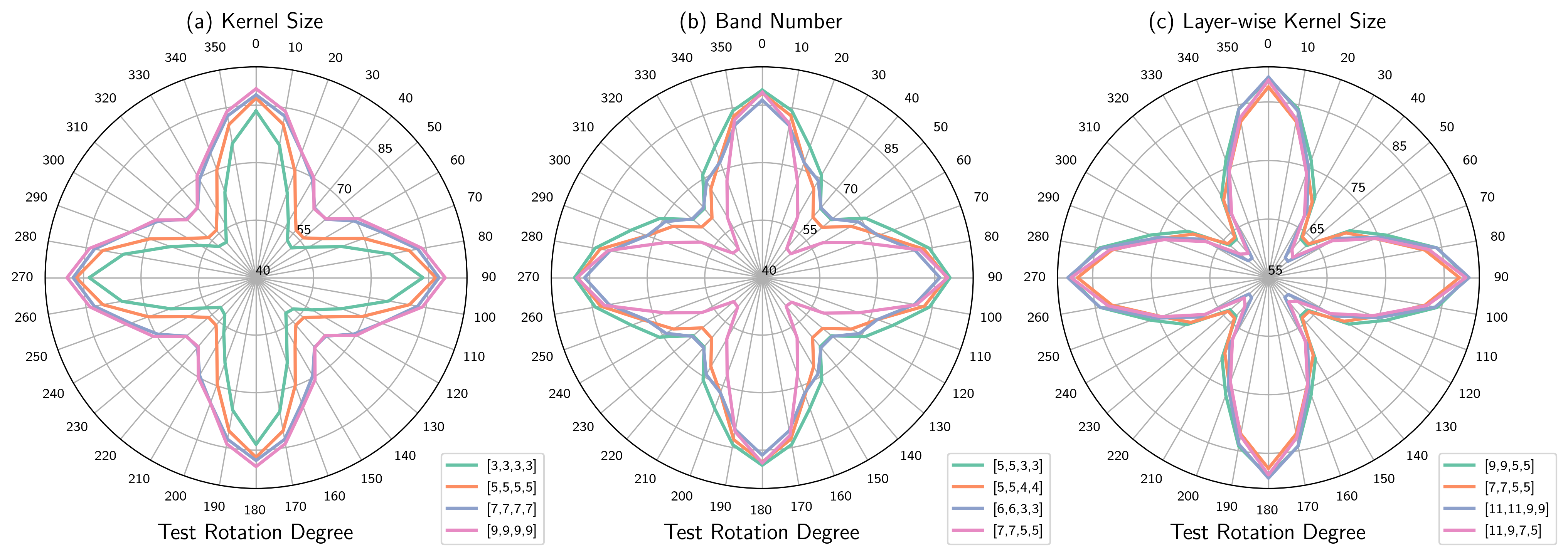}
    \caption{{\bf CIFAR-10~\cite{krizhevsky2009learning} Ablation Experiment Across Rotation Degree}. We plot the detailed accuracy curve for each rotation angle in a radar plot for each group of ablation experiments (\cref{tab:ablation1}). All the models show a similar periodic pattern. Our choice of model design shows the best performance.}  
    \label{fig:cifar_abla}
    \vspace{-3mm}
\end{figure*}

\section{Kernel Design Analysis}
\label{sec:kernel_design}

We detail the analysis for different kernel design results presented in \cref{tab:arch}. 
\Cref{fig:cifar_kdesign} plots the per-angle accuracy on the CIFAR-10~\cite{bian2022learning} dataset. From \cref{fig:cifar_kdesign}(a), we note that using the circular kernel helps to improve the model's consistency to maintain the rotational equivariance across different rotations. \Cref{fig:cifar_kdesign}(b) shows the improvement introduced by the mixture of Gaussian weighting smoothed kernel, which makes the discrete kernel closer to the continuous form and improves the performance on off-axes rotations. We note from \cref{fig:cifar_kdesign}(c) that even if the fixed $\Theta_\sigma$ can bring a better consistency across rotation, it also limits the model's capability. By making the $\Theta_\sigma$ a layer-wise trainable parameter, our model's ability to handle non-rotated images is improved.

\section{Kernel Configuration Analysis}
\label{sec:kernel_config}

\paragraph{Kernel Size} We compare different kernel sizes in the network architecture (\cref{tab:ablation1}). 
Here, we set the \convabb kernel size to be the same in all four blocks. While model size increases linearly with respect to kernel size, the performance increases correspondingly. Additionally, we note that a smaller convolutional kernel not only limits the model's absolute performance on the original test set but also results in a greater discretization error in the rotated test set. This also validated our claim in \cref{sec:intro}.

\mypar{Number of Rings} We further evaluate the influence of different ring numbers: $n$ (\cref{tab:ablation1}). Here, we fix the kernel size to be $[9, 9, 5, 5]$. We use the base ring number $n = \lfloor k/2\rfloor + 1$ with kernel size $k$. Using more rings results in a loss in performance, even if the parameter size is increased. This further proved our intuition of applying kernel smoothing. Having more rings is essentially equivalent to having a more discretized kernel, while the absolute performance on the original test set may remain relatively the same, the performance in the rotated test, especially angles like $45^{\circ}$, $135^{\circ}$, $225^{\circ}$, and $315^{\circ}$ will degenerate significantly. This, together with our observation about the kernel size in the previous experiment, reveals the reason for the failure of existing na\"{i}ve radial symmetric kernel methods~\cite{yeh2016stable,kohli2017learning,dudar2019use,fuhl2021rotated,zhang2019shellnet}. A small and highly discretized kernel will limit both the expressivity and the capability of maintaining equivariance in application. Our \convname smoothed the kernel to alleviate the discretization error. The efficient parametrization and computation strategy allows the use of larger convolutional kernels for better expressivity.

\mypar{Layer-wise Kernel Size} We compare the effect of using different kernel sizes for each layer (\cref{tab:ablation1}). 
Our choice of using smaller kernels in the deeper layer reduces the number of parameters by $\sim$30\% while achieving a top performance. Other kernel size configurations bring no improvement. 

\mypar{Per-angle Results} We detail the per-angle ablation experiment results above in \cref{fig:cifar_abla}. We note that in  \cref{fig:cifar_abla}(a), the kernel size contributes most to the overall performance difference, where a smaller kernel size tends to result in lower performance, while the number of bands and layer-wise kernel size mainly influences the performance in the ``intermediate'' angles (\cref{fig:cifar_abla}(b,c)). However, these two parameters can greatly influence the model size, which allows us to achieve a similar level of performance with a smaller number of parameters in the chosen configuration.

\begin{figure}[!t]
    \centering
    \includegraphics[width=\columnwidth]{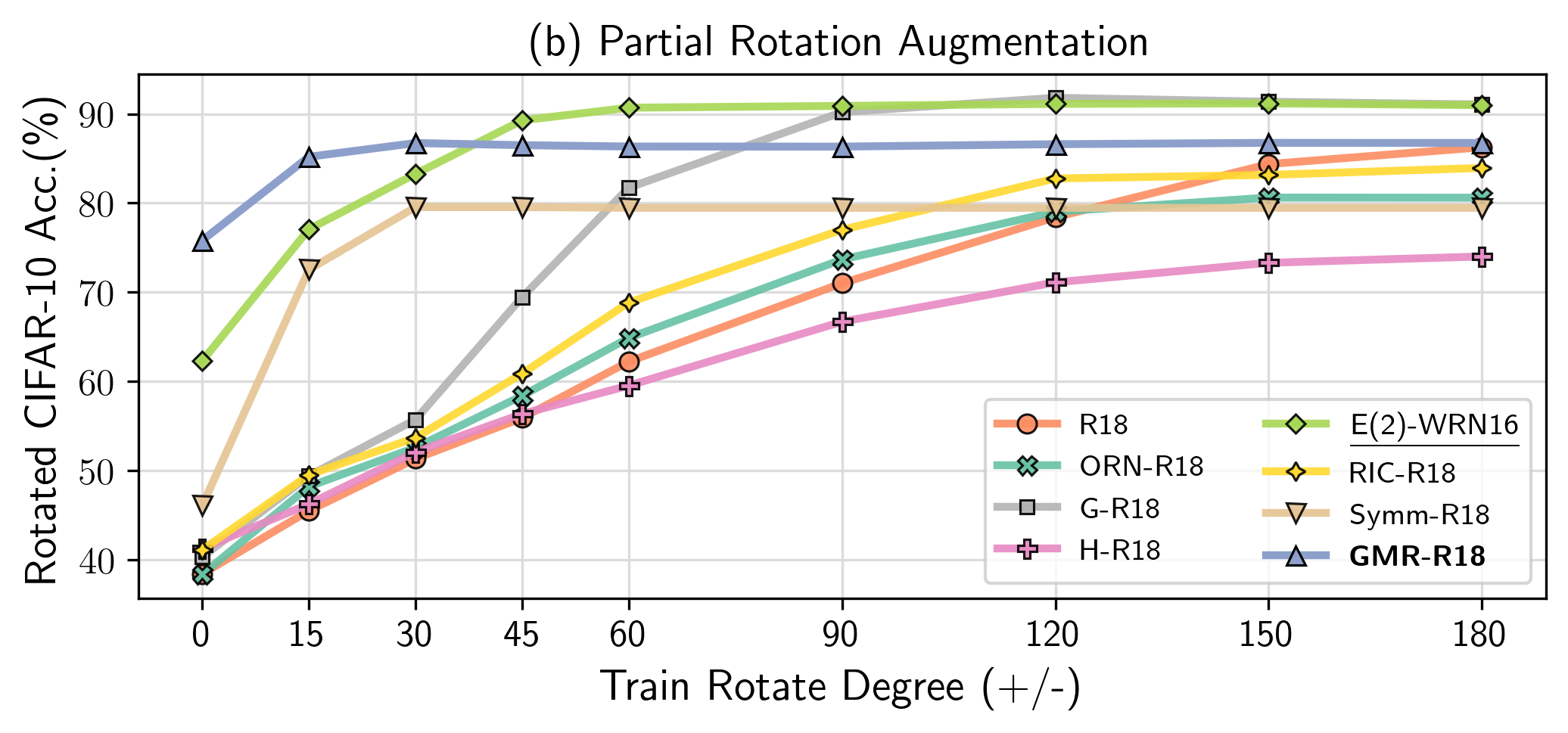}
    % \vspace{-1.7\baselineskip}
    \caption{\textbf{Evaluating the Effect of Partial Data Augmentation. } We evaluate model performance on rotated CIFAR-10 test sets under different training rotation augmentation ($\pm$) angles.}  
    \label{fig:partial}
    \vspace{-4mm}
\end{figure}  

\begin{table*}[!t]
\centering

\setlength{\tabcolsep}{4pt}
\resizebox{0.95\textwidth}{!}
{
\begin{tabular}{lcccc|ccc|ccc|ccc|ccc}
\hline
\multicolumn{1}{c}{\multirow{2}{*}{\textbf{Models}}} & \multicolumn{1}{c}{\multirow{2}{*}{\textbf{\#Param}}} & \multicolumn{3}{c|}{\textbf{CIFAR-10}~\cite{krizhevsky2009learning}} & \multicolumn{3}{c|}{\textbf{NWPU-10}~\cite{su2019object}} & \multicolumn{3}{c|}{\textbf{MTARSI}~\cite{wu2020benchmark}} & \multicolumn{3}{c|}{\textbf{NCT-CRC}~\cite{kather2019predicting}} & \multicolumn{3}{c}{\textbf{PCam}~\cite{bejnordi2017diagnostic}} \\ \cline{3-17} 
\multicolumn{1}{c}{} & \multicolumn{1}{c}{} & Orig. & Rot. & Ref. & Orig. & Rot. & Ref. & Orig. & Rot. & Ref. & Orig. & Rot. & Ref. & Orig. & Rot. & Ref. \\ \hline
ResNet-50~\cite{liu2022convnet} & 23.5M & \textbf{94.7} & 37.1 & 65.7 & 96.5 & 88.9 & 95.2 & 92.4 & 57.2 & 64.9 & 92.9 & 88.5 & 92.4 & \textbf{86.1} & 75.2 & 83.7 \\ \rowcolor[HTML]{EFEFEF}
\convabb-ResNet50 & 16.1M & 89.7 & \textbf{80.9} & \textbf{89.8} & \textbf{96.8} & \textbf{96.4} & \textbf{96.8} & \textbf{96.4} & \textbf{92.9} & \textbf{96.3} & \textbf{95.0} & \textbf{94.1} & \textbf{95.0} & 85.8 & \textbf{84.9} & \textbf{84.9} \\ \hline
ConvNeXt-T~\cite{liu2022convnet} & 27.8M & \textbf{92.1} & 36.8 & 62.3 & \textbf{93.9} & 86.5 & 91.9 & \textbf{88.3} & 53.2 & 62.0 & \textbf{94.0} & 78.7 & 91.4 & 71.5 & 61.9 & 69.5 \\ \rowcolor[HTML]{EFEFEF}
\convabb-ConvNeXt-T & 26.3M & 87.7 & \textbf{64.9} & \textbf{87.7} & 91.6 & \textbf{93.3} & \textbf{92.3}& 82.6 & \textbf{82.8} & \textbf{82.6} & 92.3 & \textbf{91.5} & \textbf{92.3} & \textbf{72.4} & \textbf{72.3} & \textbf{72.4} \\ \hline
DenseNet-121~\cite{huang2017densely} & 7.0M & 87.9 & 34.7 & 60.0 & \textbf{97.8} & 93.1 & \textbf{95.9} & \textbf{91.5} & 55.7 & 64.4 & \textbf{94.7} & 84.6 & 93.1 & \textbf{83.9} & 71.6 & 80.8 \\ \rowcolor[HTML]{EFEFEF}
\convabb-DenseNet-121 & 5.7M & \textbf{88.6} & \textbf{69.1} & \textbf{88.6} & 95.1& \textbf{95.4} & 95.1& 88.7 & \textbf{88.0} & \textbf{88.7} & 94.1 & \textbf{93.7} & \textbf{94.1} & 81.1 & \textbf{80.7} & \textbf{81.1} \\ \hline
DenseNet-161~\cite{huang2017densely} & 26.5M & 88.5 & 36.0 & 61.3 & \textbf{97.3} & 92.4 & \textbf{96.7} & \textbf{91.6} & 55.6 & 64.2 & 93.9 & 85.9 & 92.4 & \textbf{84.3} & 72.8 & 82.1 \\ \rowcolor[HTML]{EFEFEF} 
\convabb-DenseNet-161 & 22.5M & \textbf{89.5} & \textbf{71.3} & \textbf{89.5} & 94.2& \textbf{94.7} & 94.1 & 87.5 & \textbf{87.5} & \textbf{87.7} & \textbf{94.4} & \textbf{94.0} & \textbf{94.4} & 83.5 & \textbf{83.8} & \textbf{83.5} \\ \hline
\end{tabular}
}
\caption{\textbf{Evaluation of \convname on Other Model Architectures.} We present the accuracy of each baseline on the CIFAR-10~\cite{krizhevsky2009learning}, NWPU-10~\cite{su2019object}, MTARSI~\cite{wu2020benchmark}, NCT-CRC-100k~\cite{kather2019predicting}, and PatchCamelyon~\cite{bejnordi2017diagnostic} datasets with larger ResNet~\cite{he2016deep}, ConvNeXt~\cite{liu2022convnet}, and DenseNet~\cite{huang2017densely}. We train each model without rotation and evaluate the test set without rotation (Orig.), with rotations (Rot.), and with reflection (Ref.). We report the average accuracy of each test set. We highlight the best performance in bold. Our model is shaded in gray.}
\label{tab:other_arch}
\vspace{-2mm}
\end{table*}

\section{Different Model Architectures}
\label{sec:arch}

The proposed \convname layer is designed to be \textit{plug-and-play} so it can be easily integrated into different CNN architectures. To evaluate the effectiveness of our \convname in various architectures, we compare the adapted \convabb-ConvNeXt and \convabb-DenseNet with original models~\cite{liu2022convnet,huang2017densely} following our rules for Equivariant CNN (\cref{sec:rules}) and \cref{sec:implement} on all five datasets (\cref{tab:other_arch}). Our model shows a consistent and notable improvement on the rotated test set for all four model architectures across the five datasets. Furthermore, our \convname models also improve the performance of the reflected test set in most cases. Although the symmetric constraint may affect our model's expressibility, our model still achieves a comparable performance on the original test set.

\section{Evaluating Partial Rotation Augmentation}
\label{sec:partial}

We evaluate the effect of rotation augmentation during training on CIFAR-10 for all models.
We apply training rotation augmentation with different maximum-allowed rotation angles to each model (\cref{fig:partial}). 
Our model is the fastest to saturate with partial rotation augmentation of only $[-15^\circ,15^\circ]$. This illustrates how our method will succeed when training augmentation is insufficient or cannot cover all scenarios that will appear in the test set. 
Symm-R18~\cite{dudar2019use} saturates next at 30 degrees, but its optimal performance is 5\% lower than our \convabb-R18. 
Meanwhile, E(2)-WRN16\cite{weiler2019general} and G-R18~\cite{cohen2016group} achieve higher overall performance with sufficient augmentation but saturate more slowly. 
Considering the extra size and computational cost, it is not unexpected that they have higher accuracy when full augmentation is applied.

\begin{figure}[t!]
    \centering
    \includegraphics[width=0.9\columnwidth]{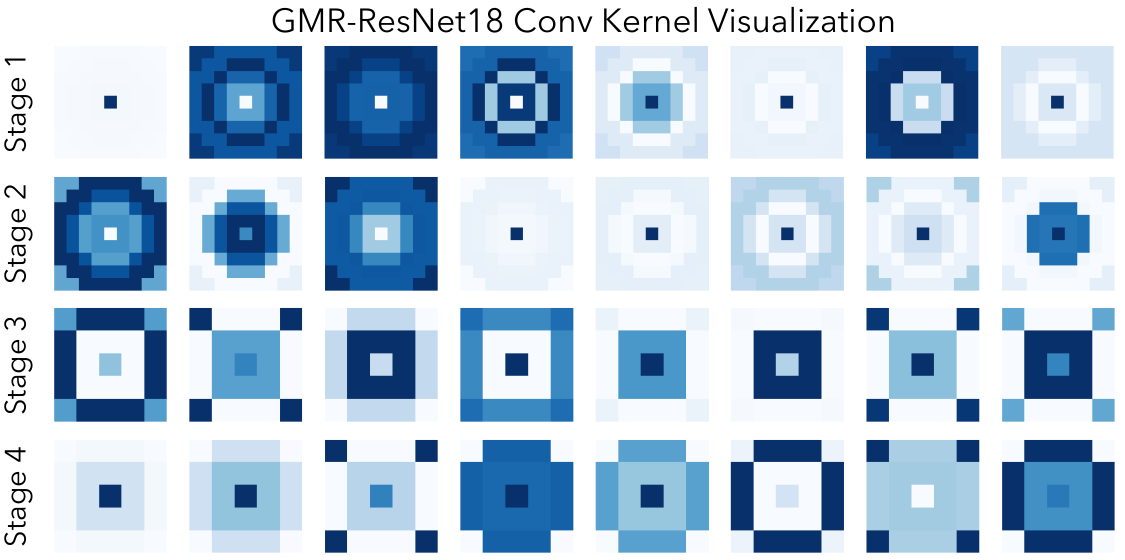}
    \caption{{\bf \convabb-R18 Kernel Visualization}. We randomly select 8 kernels at each stage of our \convabb-R18 model trained with the CIFAR-10 dataset and plot these kernels. The first two stages have a size of 9, and the latter two stages have a size of 5. The corners of these kernels are zero, but they have different scales, resulting in a different corner color.}
    \label{fig:sri_kernel}
    \vspace{-3mm}
\end{figure}  

\begin{figure}[!t]
    \centering
    \includegraphics[width=1.0\columnwidth]{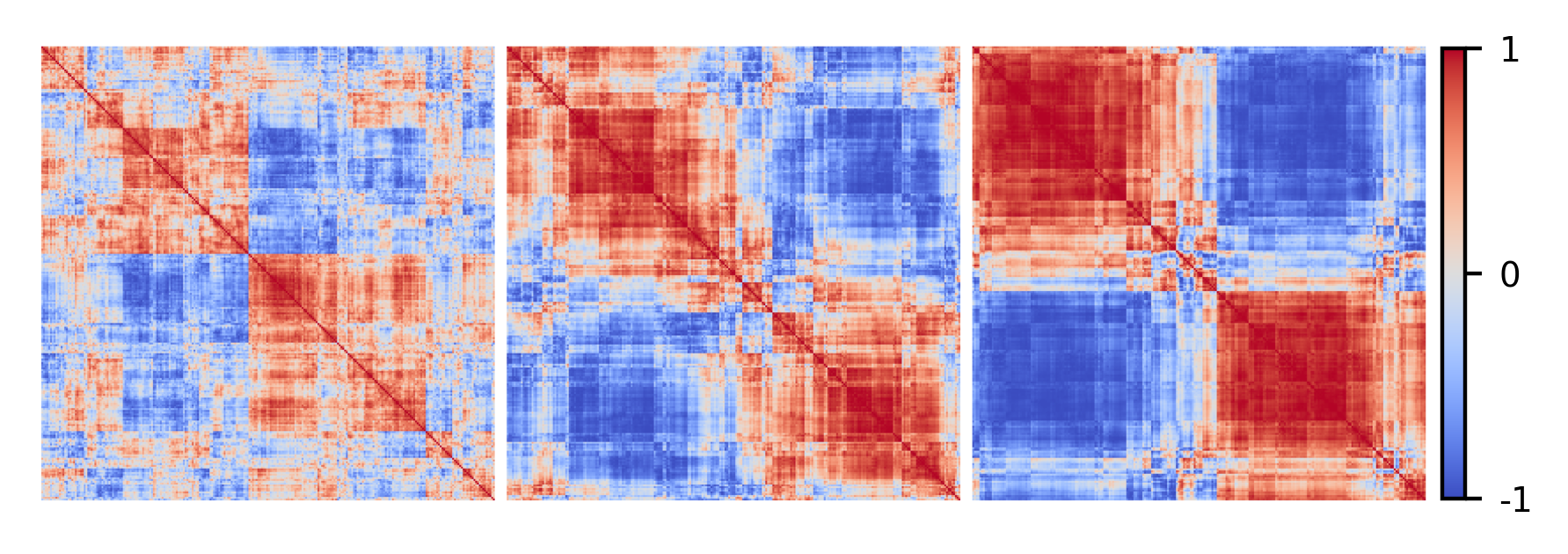}
    \caption{{\bf Correlation Map for Conv.~Kernel Weight}. After training the model on the CIFAR-10~\cite{krizhevsky2009learning}, we plot the cosine correlation map between each convolutional kernel for the conventional convolution layer in R18 (left) and our \convname layer without (center) and with Gaussian Mixture Ring (right). Red regions indicate higher positive correlations between individual kernels while blue regions indicate stronger negative correlations.}  
    \vspace{-2mm}
    \label{fig:corr}
\end{figure}  

\begin{figure*}[t!]
    \centering
    \includegraphics[width=\textwidth]{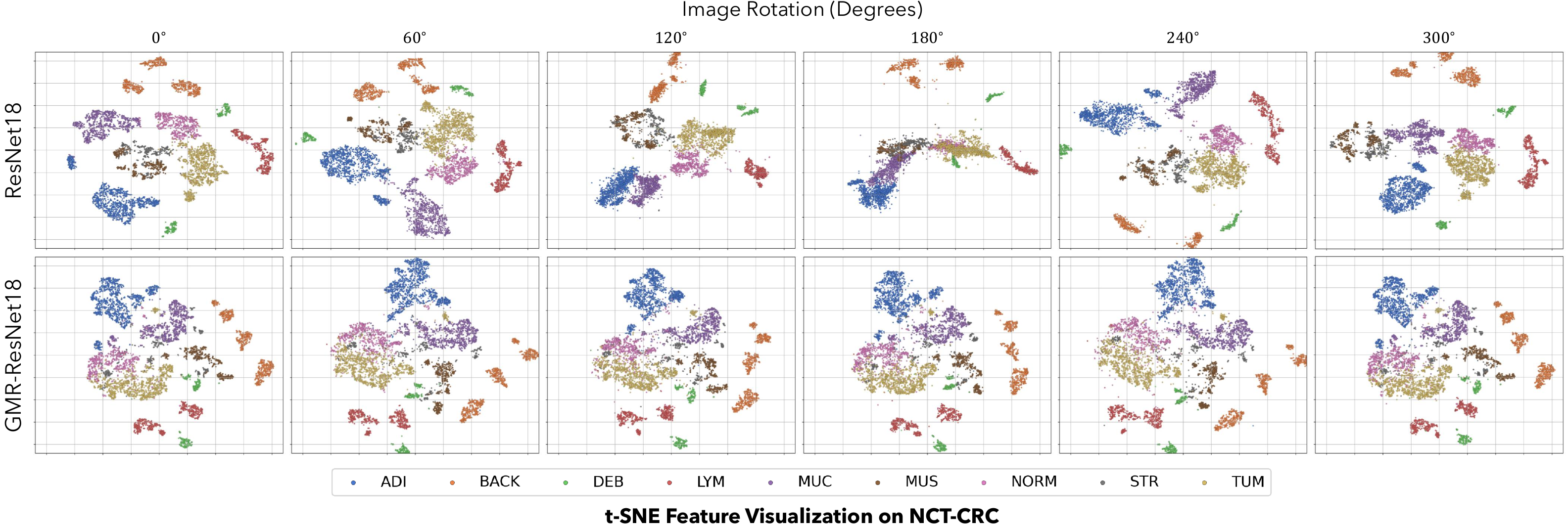}
    \caption{{\bf t-SNE Visualization on NCT-CRC Test Set.} We visualize the clustered test set samples in the NCT-CRC~\cite{kather2019predicting} dataset using t-SNE. The test input is rotated before feeding into each model. We compare the visualization between vanilla ResNet18~\cite{he2016deep} and our \convabb-ResNet18 trained with no rotational augmentation. We colorize the samples in the clustered results according to their classification label, as shown in the legend.}
    \label{fig:cluster}  \vspace{-3mm}
\end{figure*}  

\section{Visualizing Learned GMR Kernels}

We visualize randomly selected kernels from our \convabb-R18 model trained on the CIFAR-10 dataset in \cref{fig:sri_kernel}. Our \convname is strictly symmetric with a fixed number of rings. As the kernel size increases, the rings better approximate a circular shape, and the rotation-equivariant property is stronger. Also, we note that the parameters between each ring are roughly on the same scale and are changing smoothly. This is mainly because of kernel smoothing with a mixture of Gaussian weighting functions.

We further evaluate the amount of redundancy in the trained kernels by comparing the correlation of kernel parameters within a given level for both the conventional R18 and \convabb-R18 trained on the CIFAR-10~\cite{krizhevsky2009learning} dataset (\cref{fig:corr}). We choose the first convolutional layer in the second stage of the R18 model (128 features), where our model has a kernel size of 9 and ring number of 5, which has a similar number of trainable parameters within each kernel compared to standard convolutional kernels of size 3. We compute the cosine correlation between each convolutional kernel and visualize the correlation in \cref{fig:corr}. When computing the cosine correlation for our model, we compute the value only for the trainable parameter tensor instead of the full 9$\times$9 convolutional kernel. We note that our \convname has a more polarized correlation, \ie, either highly positively or highly negatively correlated. This pattern is more obvious when we apply the Gaussian ring smoothing, which is reasonable since the values of each kernel are smoothed. However, the traditional convolutional kernel tends to have more kernels that are orthogonal to each other. A similar pattern is observed in other convolutional layers. This indicates that there is substantial information redundancy within our model, which hints that our model has the potential to further reduce its number of parameters.

\begin{figure*}[t!]
    \centering
    \includegraphics[width=0.9\textwidth]{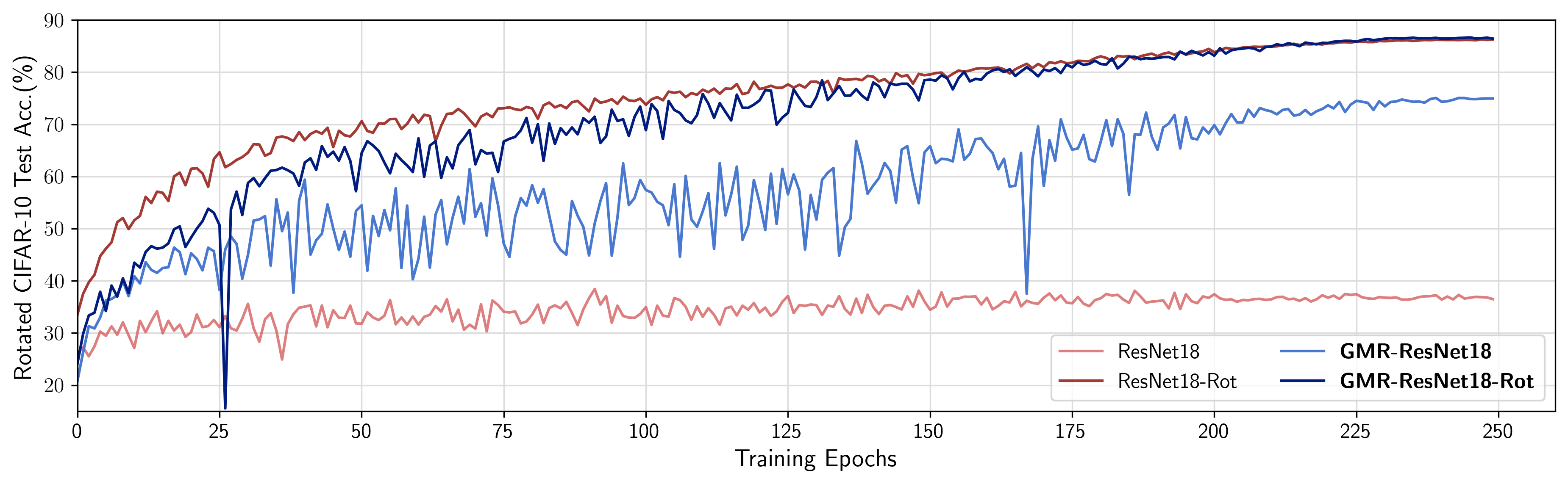}
    \caption{{\bf CIFAR-10~\cite{krizhevsky2009learning} Rotated Test Accuracy During Training}. We plot averaged accuracy on the rotated CIFAR-10 test set for each training epoch during training for standard ResNet18 and our equivariant \convabb-ResNet18. ``Rot'' indicates the model is trained with full random rotation $[-180\text{\textdegree}, 180\text{\textdegree}]$
    augmentation, which is plotted with deeper color.}  
    \label{fig:cifar_train}
    % \vspace{-3mm}
\end{figure*}

\section{Robust Equivariant Feature Embeddings}
\label{sec:eq_feat}

Using the models trained on the NCT-CRC~\cite{kather2019predicting} dataset classification task, we demonstrate the capability of \convabb-CNN to learn equivariant imaging feature embeddings by removing the final classification layers of the R18 and \convabb-R18 architectures.
We then performed spatial dimensional average pooling of feature maps from the final convolution layer to produce a single vector representation of a given input image.
We extract the feature embeddings from the trained R18 and \convabb-R18 models for inputting all NCT-CRC testing set images rotated at 60{\textdegree} increments.
To visualize the consistency of feature map embeddings across rotated images, we perform unsupervised non-linear dimensionality reduction using t-distributed Stochastic Neighbor Embedding (t-SNE)~\cite{van2008visualizing}.
For each model, embeddings across all rotation angles were used as input for the t-SNE mapping into a two-dimensional space for qualitative visual assessment.
\Cref{fig:cluster} visualizes the embeddings in t-SNE space by rotation angle with points color-coded by ground-truth classification label. 
Features that are equivariant to rotation should not move substantially in t-SNE space when the input image is rotated.
The standard R18 feature embeddings move substantially within the t-SNE space as the images are rotated, and we observe labeled clusters mixing as the images rotate.
In contrast, the \convabb-CNN feature embeddings remain remarkably stable, and labeled clusters remain mostly well-separated.
These results demonstrate both the stability of equivariant feature embeddings using our \convabb-CNN and the instability of non-equivariant feature embeddings.
The consistent performance of features is critical for developing robust imaging representations.

\section{Feature Map Analysis}

\begin{figure*}[t!]
    \centering
    \includegraphics[width=0.78\textwidth]{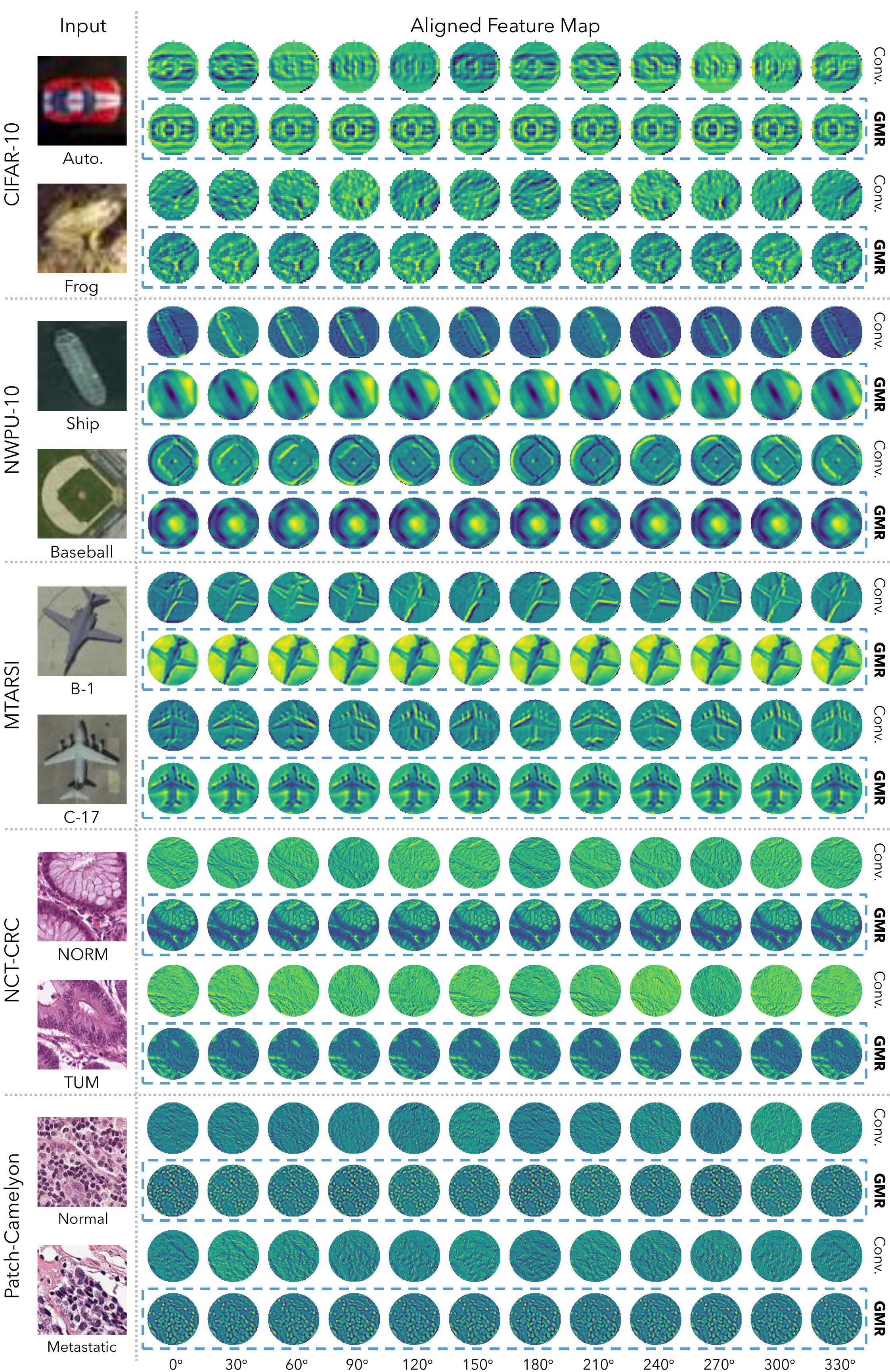}
    \caption{{\bf Feature Map Visualization.} We visualize the feature map for conventional CNN (``Conv.'' rows) and our \convabb-CNN (``\convabb'' rows) for different rotation angles on all five datasets. The visualized feature map is rotated back to align with the original orientation. Our method shows a more uniform and symmetric behavior across 12 rotation angles. We highlight our results with a blue boundary.}  
    \label{fig:feat}  \vspace{-3mm}
\end{figure*}

We visualize feature maps for the R18 and \convabb-R18 trained on the five validation datasets in \cref{fig:feat}. We rotate the same input image by 12 different angles and extract the feature map from the first convolutional layer in the first stage for higher resolution and interpretability. We rotate the feature map back to align with the original orientation using the inverse rotation and mask the central region with a circle mask to avoid visualizing border areas affected by image padding. We note that our \convname performs more uniformly across different angles, and there is a strict periodic pattern every 90\textdegree~in our \convname. This illustrates the rotational equivariance in our method. We further highlight the result from the orientation-independent NWPU-10~\cite{su2019object}, MTARSI~\cite{wu2020benchmark}, NCT-CRC~\cite{kather2019predicting}, and Patch-Camelyon~\cite{bejnordi2017diagnostic}, where the image is taken \emph{above} the object, and the orientation of input is arbitrary. Among these few examples, our model is able to generate more uniform and consistent features across rotations, while the features of the conventional convolutional layer change with different rotations.

\section{Model Convergence Analysis}

We further evaluate the convergence of our model on unseen rotations during training. We compare our \convabb-R18 and conventional R18 models' performance on the rotated CIFAR-10~\cite{krizhevsky2009learning} test set for every epoch during training (\cref{fig:cifar_train}). We also introduce the full random $[-180\text{\textdegree}, 180\text{\textdegree}]$ training augmentation for each model. We note that our model performs consistently better than the conventional CNN when there is no data augmentation. However, we also noticed that our model has a test performance with larger oscillations, which can be introduced by the strong symmetric constraint. Meanwhile, when the full augmentation is applied, our model can achieve a similar level of performance with only 34\% of parameters throughout the whole training process. The oscillation during training is also smaller when augmentation is applied.

\section{Alternative Rotated Performance Figures}

\begin{figure*}[t!]
    \centering
    \includegraphics[width=0.95\textwidth]{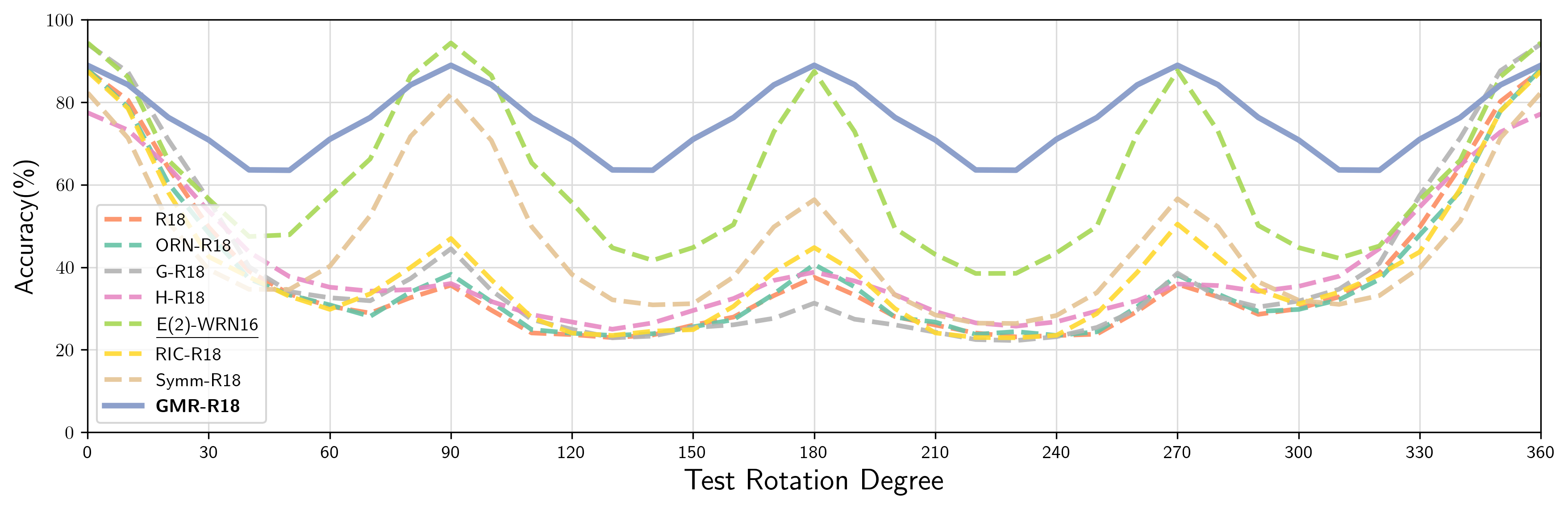}
    \caption{{\bf CIFAR-10~\cite{krizhevsky2009learning} Accuracy Across Rotation Degree}. We plot the same accuracy curve for each rotation degree in a line plot as in \cref{fig:radar3}(a). Our model outperforms all the other baselines in this evaluation and it shows much less performance degradation when an intermediate rotation degree is present.}  
    \label{fig:cifar_line}
    \vspace{-3mm}
\end{figure*}

\begin{figure*}[t!]
    \centering
    \includegraphics[width=0.95\textwidth]{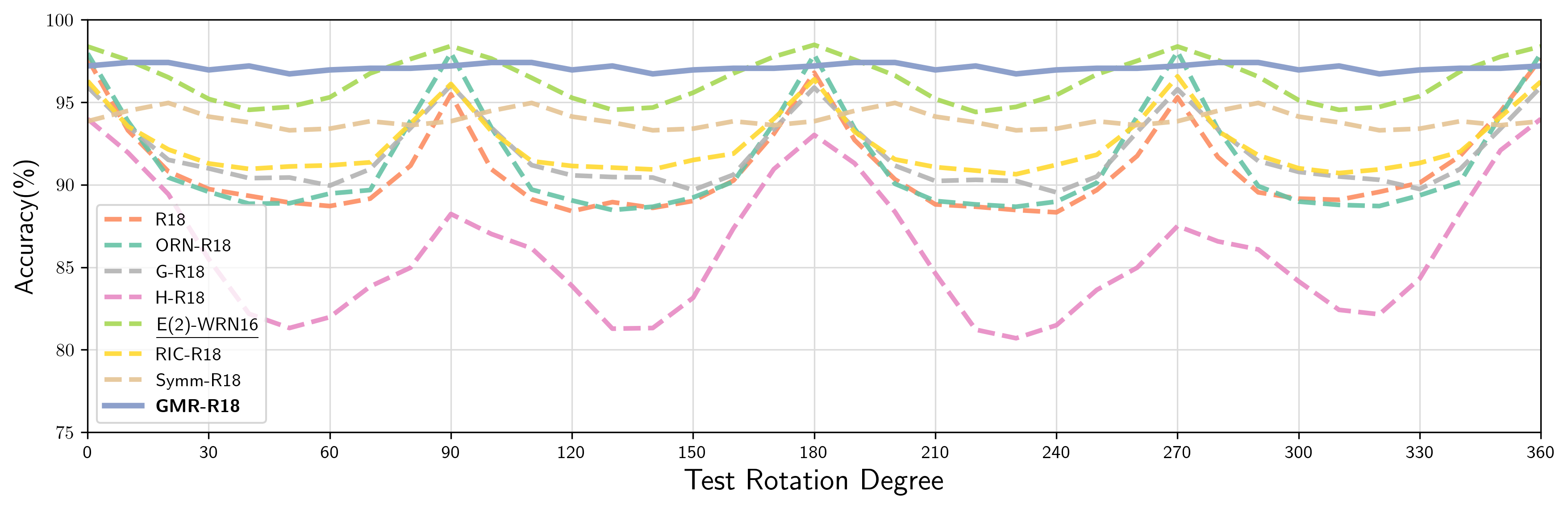}
    \caption{{\bf NWPU-10~\cite{su2019object} Accuracy Across Rotation Degree}. We plot the same accuracy curve for each rotation degree in a line plot as in \cref{fig:radar3}(b).}  
    \label{fig:vhr_line}
    \vspace{-3mm}
\end{figure*}

\begin{figure*}[t!]
    \centering
    \includegraphics[width=0.95\textwidth]{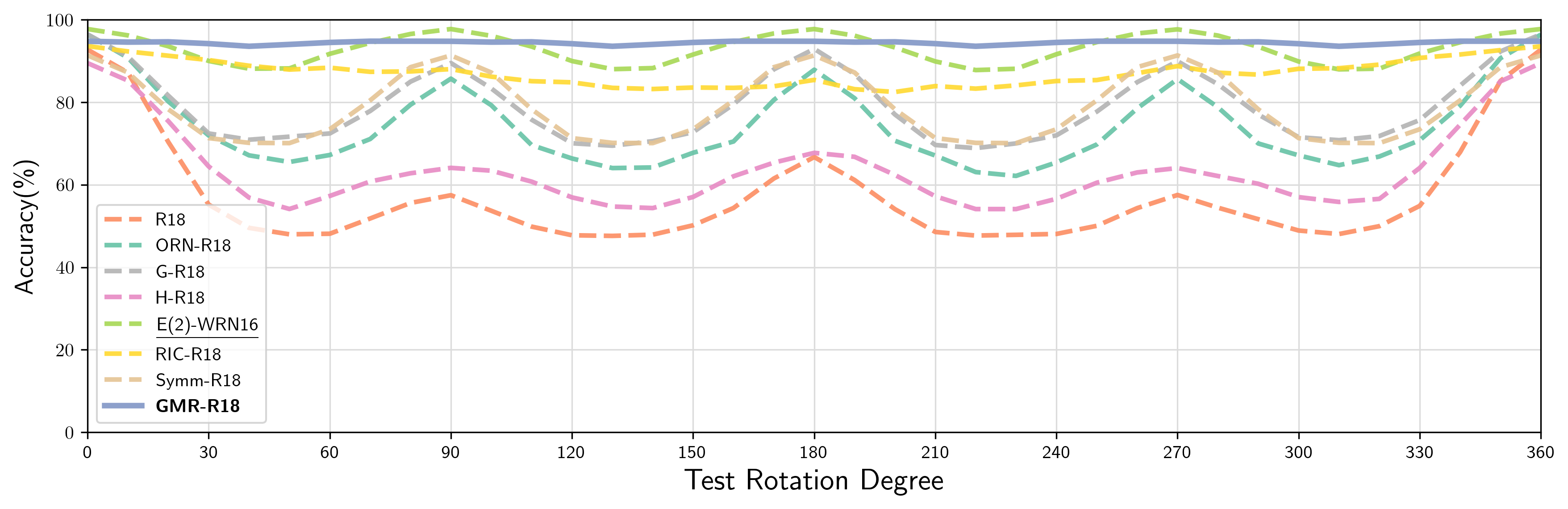}
    \caption{{\bf MTARSI~\cite{wu2020benchmark} Accuracy Across Rotation Degree}. We plot the same accuracy curve for each rotation degree in a line plot as in \cref{fig:radar3}(c).}  
    \label{fig:mtarsi_line}
    \vspace{-3mm}
\end{figure*}

\begin{figure*}[t!]
    \centering
    \includegraphics[width=0.95\textwidth]{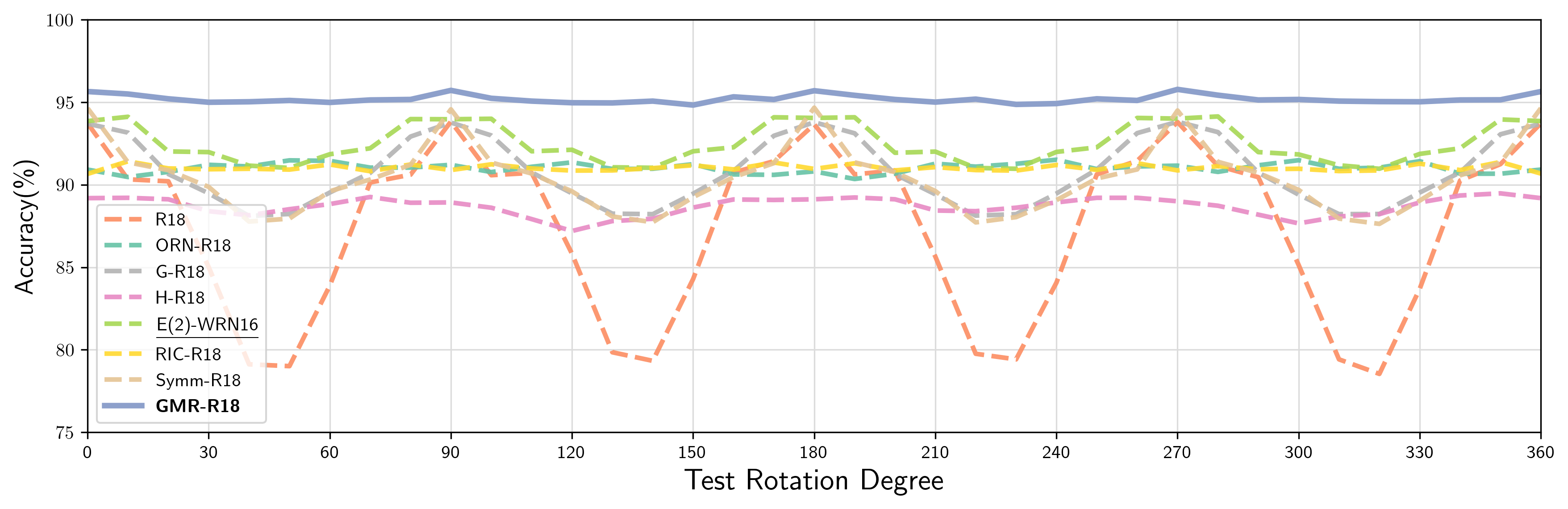}
    \caption{{\bf NCT-CRC~\cite{kather2019predicting} Accuracy Across Rotation Degree}. We plot the same accuracy curve for each rotation degree in a line plot as in \cref{fig:radar3}(d).}  
    \label{fig:crc_line}
    \vspace{-3mm}
\end{figure*}

\begin{figure*}[t!]
    \centering
    \includegraphics[width=0.95\textwidth]{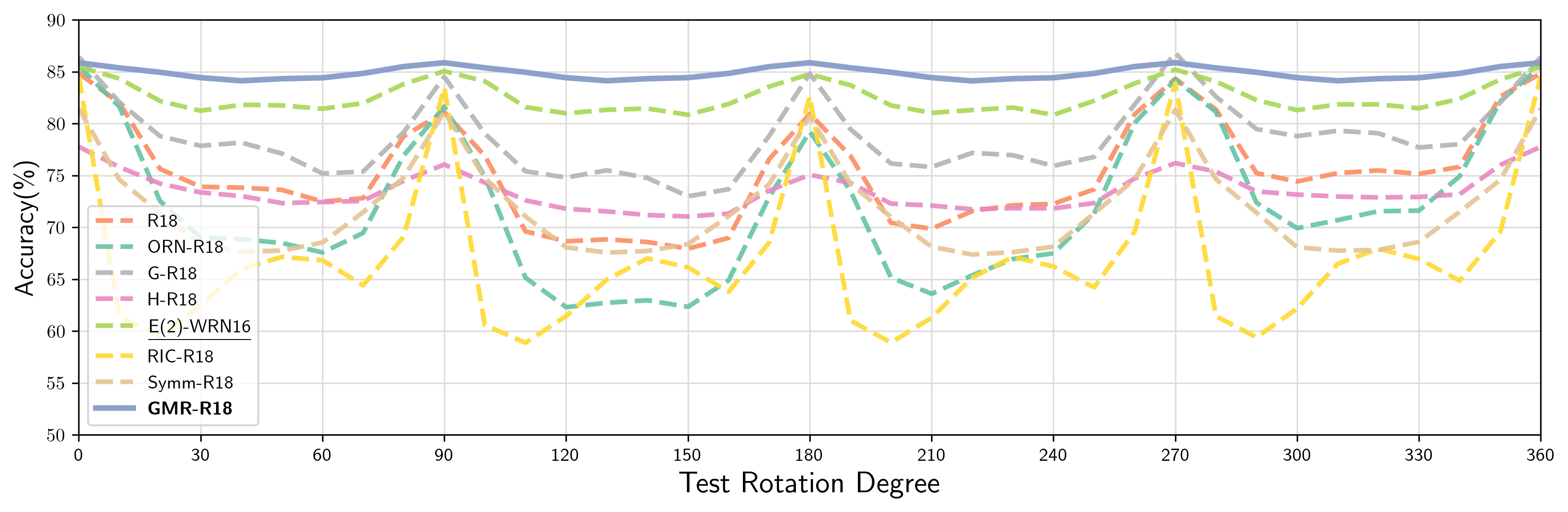}
    \caption{{\bf Patch-Camelyon~\cite{krizhevsky2009learning} Accuracy Across Rotation Degree}. We plot the same accuracy curve for each rotation degree in a line plot as in \cref{fig:radar3}(e). Our model outperforms all the other baselines in this evaluation, and it shows much less performance degradation when an intermediate rotation degree is present.}  
    \label{fig:pcam_line}
    \vspace{-3mm}
\end{figure*}

As an alternative visualization of performance results, we plot the exact same figures shown in \cref{fig:radar3}, as line plots in 
\cref{fig:cifar_line,fig:mtarsi_line,fig:vhr_line,fig:crc_line,fig:pcam_line} 
respectively, for all five validation datasets.

\clearpage

\end{document}